\documentclass[preprint,12pt]{elsarticle}

\usepackage[numbers]{natbib}

\usepackage{amsmath}
\usepackage{graphicx}
\usepackage{algorithm}
\usepackage{algorithmic}
\usepackage{multirow}
\usepackage{standalone}
\usepackage[table,xcdraw]{xcolor}
\usepackage{float}
\usepackage{amssymb}
\usepackage{hyperref}
\usepackage{lscape}

\def\tsc#1{\csdef{#1}{\textsc{\lowercase{#1}}\xspace}}
\tsc{WGM}
\tsc{QE}
\tsc{EP}
\tsc{PMS}
\tsc{BEC}
\tsc{DE}

\begin{document}
\let\WriteBookmarks\relax
\def\floatpagepagefraction{1}
\def\textpagefraction{.001}



\title{No-reference based automatic parameter optimization for iterative reconstruction using a novel search space aware crow search algorithm}

%
\author[1,2]{Poorya MohammadiNasab}


\affiliation[1]{organization={Department of Medicine},
    addressline={Danube Private University (DPU)}, 
    city={Krems},
    country={Austria}}

\affiliation[2]{organization={Center for Medical Physics and Biomedical Engineering},
    addressline={Medical University Vienna}, 
    city={Vienna},
    country={Austria}}

\author[3]{Ander Biguri}

\affiliation[3]{organization={Department of Applied Mathematics and Theoretical Physics},
    addressline={University of Cambridge}, 
    city={Cambridge},
    country={United Kingdom}}
    
\author[4]{Philipp Steininger}
\author[4]{Peter Keuschnigg}
\author[4]{Lukas Lamminger}
\author[4]{Agnieszka Lach}


\affiliation[4]{organization={MedPhoton GmbH},
    addressline={Salzburg}, 
    country={Austria}}


\author[1,2,5]{S M Ragib Shahriar Islam}

\author[2,3]{Anna Breger}

\author[1,2,6]{Clemens Karner}

\author[3]{Carola-Bibiane Schönlieb}

\author[2]{Wolfgang Birkfellner}

\author[1,5,2]{Sepideh Hatamikia\corref{cor1}}
\ead{Sepideh.Hatamikia@dp-uni.ac.at}

\affiliation[6]{organization={Faculty of Mathematics},
    addressline={University of Vienna}, 
    city={Vienna},
    country={Austria}}

\affiliation[5]{organization={Austrian Center for Medical Innovation and Technology (ACMIT)},
    addressline={Wiener Neustadt}, 
    country={Austria}}

\cortext[cor1]{Corresponding author}



\begin{abstract}
Unlike traditional back-projection methods, which require a high number of projections to reconstruct images, iterative reconstruction techniques deliver substantially better results in limited-projection scenarios. Consequently, their ability to reduce radiation exposure by using fewer projections has attracted significant attention.
However, these methods typically require a precise tuning of several hyperparameters, which can have a major impact on reconstruction quality. Manually setting these parameters is time-consuming and increases the workload for human operators. In this paper, we introduce a novel fully automatic parameter optimization framework that can be applied to a wide range of Cone-beam computed tomography (CBCT) iterative reconstruction algorithms to determine optimal parameters without requiring a reference reconstruction. The proposed method incorporates a modified crow search algorithm (CSA) featuring a superior set–dependent local search mechanism, a search-space-aware global search strategy, and an objective-driven balance between local and global search. Additionally, to ensure an effective initial population, we propose a chaotic diagonal linear uniform initialization scheme that accelerates algorithm convergence. The performance of the proposed framework was evaluated on three imaging machines and four real datasets, as well as three different iterative reconstruction methods with the highest number of tunable parameters, representing the most challenging senario. The results indicate that the proposed method could outperform manual settings and CSA, with an 4.19\% improvement in average fitness and 4.89\% and 3.82\% improvements on CHILL@UK and RPI\_AXIS, respectively, which are two benchmark no-reference learning-based quality metrics. In addition, the qualitative results clearly show the superiority of the proposed method by maintaining fine details sharply. The overall performance of the proposed framework across different comparison scenarios demonstrates its effectiveness and robustness across all cases.

\end{abstract}







\maketitle

\section{Introduction}

Medical imaging is key to the advancements of modern medicine \cite{birkfellner2024applied}. X-ray computed tomography (CT) is one of the main imaging techniques used for both diagnostic and treatment purposes \cite{kalender2011computed}. Cone-beam CT (CBCT), a specialized form of CT, was originally developed for dental and orthodontic diagnostics \cite{walter2016cone, hodges2013impact, islam2024source}. Over the past years, however, its use has expanded far beyond its initial scope. CBCT has become a vital imaging modality in modern clinical practice, particularly for image-guided radiation therapy (IGRT) \cite{song2025ultra, zhou2025novel, hatamikia2021toward}, where it enables accurate patient positioning, verification of target localization, and adaptive treatment strategies. In addition, CBCT is increasingly applied in a variety of image-guided therapeutic procedures, such as interventional radiology and surgical navigation, due to its ability to provide high-resolution 3D imaging directly in the treatment room with relatively lower radiation dose and faster scan time compared to conventional CT \cite{raz2023principles}. During the imaging process, the scanner collects X-ray projections from multiple angles. These projections represent raw data and cannot be directly interpreted. Image reconstruction is the computational process that converts this raw X-ray data into cross-sectional images of internal structures.

The chosen image reconstruction technique has a direct effect on image quality, noise level, and artifacts. The Feldkamp-Davis-Kress (FDK) algorithm \cite{feldkamp1984practical} is the traditional CBCT reconstruction method, which requires a very high number of projections. Although CBCT delivers lower levels of X-ray radiation compared to conventional CT, its radiation dose still remains a major concern, as highlighted in numerous studies \cite{hatamikia2020optimization,cocskun2025awareness,bawazeer2025advancing}. Therefore, it is crucial to reduce the radiation dose by reducing the number of projections, limiting angular coverage, and lowering x-ray energy. The FDK algorithm is not well-suited for sparse-view reconstruction, as it tends to produce substantial image noise and pronounced artifacts when only a limited number of projections are available.

Iterative reconstruction methods are introduced to approach the reconstruction problem from a different perspective, and they demonstrate the potential to produce high-quality results even in limited and low-dose projection scenarios \cite{stiller2018basics, willemink2013iterative}. Although iterative reconstruction methods require higher demand for computation, this is now feasible with the advancements of faster and parallel computing hardware \cite{biguri2016tigre, hansen2018air}. In these methods, the image is refined through a repeated cycle of forward and backward projection processes. They begin with an initial random guess or a simple FDK reconstruction to simulate how the current image would appear when forward projected and then compare the simulated projections with the actual measured data. The difference between them is back projected as a projection error to update the image estimate. Regularization terms or prior models are often incorporated during the updates to stabilize the solution and reduce noise and artifacts.

\begin{figure}[]
	\centering
		\includegraphics[width=1\textwidth]{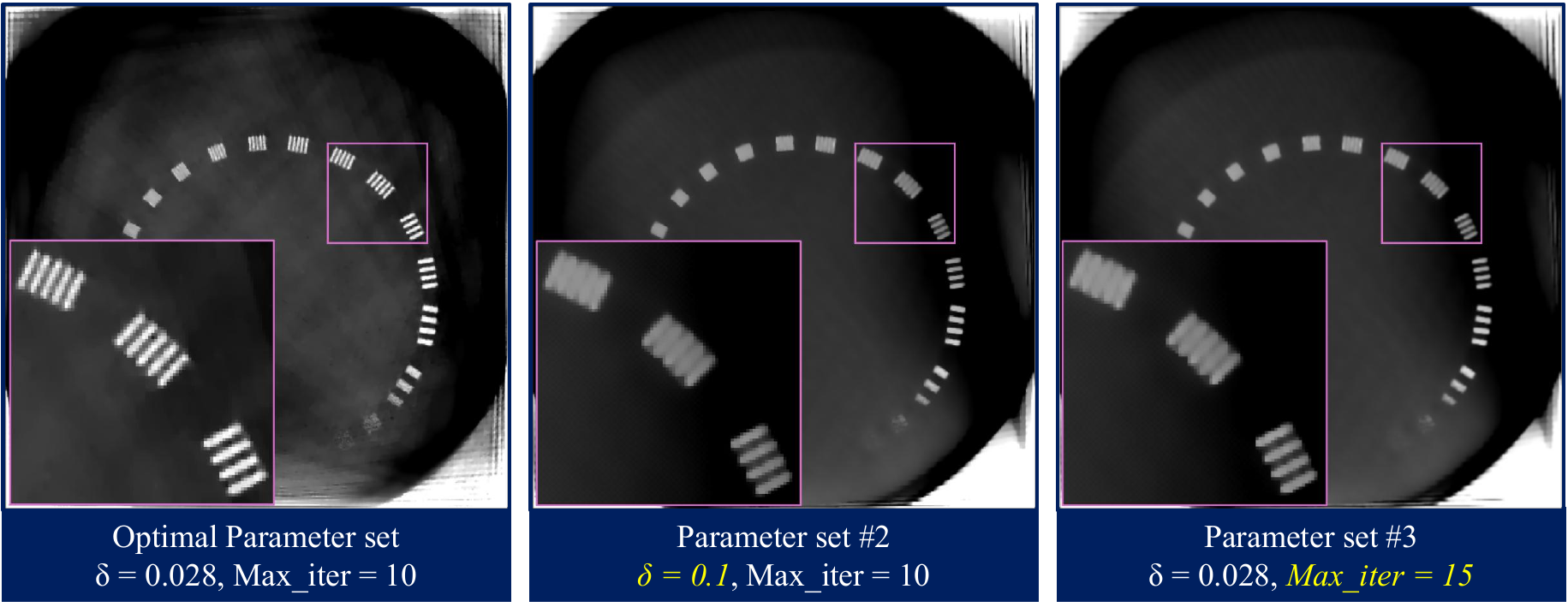}
	\caption{Reconstruction results of (A) optimized parameter set and (B and C) small variation of optimized parameters in AwPCSD algorithm.}
	\label{FIG:11}
\end{figure}

Despite the advantages of iterative reconstruction methods, they require the careful selection of hyperparameters to control the reconstruction output. In many cases, even small changes in these hyperparameters can significantly affect the final image quality as shown in figure \ref{FIG:11}. Manually tuning them is often tedious and time-consuming, particularly when multiple parameters need adjustment for a single algorithm. Furthermore, changes in imaging conditions such as switching machines, targeting different organs, or altering radiation settings typically necessitate re-tuning of the hyperparameters. Manual adjustment is also prone to observer variability, with different users achieving inconsistent performance. Automated optimization of hyperparameter sets can therefore be of great value as it saves time, reduces variability, supports process standardization, and ensures consistent reconstruction performance independent of the individual operator.

Several studies have attempted to address automatic parameter optimization for reconstruction in CT and CBCT modalities. These studies can be categorized into brute-force \cite{lohvithee2017parameter}, statistical-based \cite{duan2019region}, metaheuristics \cite{zheng2013dqs}, and more recent approaches, such as deep learning \cite{xu2021patient} and reinforcement learning \cite{shen2018intelligent}. Despite all efforts in this field, there are two notable gaps in the literature that motivate this work. First, most of the studies have focused only on optimization of one parameter \cite{shen2018intelligent, shen2019quality}, mainly  the regularization parameter, which controls the balance between the data fidelity and regularization part of the algorithm. Therefore, the development of a fully automatic parameter optimization pipeline for tuning a set of parameters has not yet been explored efficiently for algorithms such as the adaptive steepest descent projection onto convex sets (ASD-POCS) \cite{sidky2008image}, which involves eight tunable hyperparameters. Second, most existing approaches optimize the problem using a reference-based method \cite{lohvithee2017parameter, xu2011using, lohvithee2021ant}, assuming the availability of a reference reconstruction. This assumption makes them impractical in many real clinical scenarios, where ground truth data is not available.

For tuning more than one parameter of an algorithm where each parameter needs very precise tuning across a wide range of available values, there would be a huge search space, making it difficult to find the optimal point. Bio-inspired metaheuristic algorithms, on the other hand, have demonstrated good potential in exploring high-dimensional search spaces \cite{kar2016bio}. However, all metaheuristic-based methods for parameter optimization share two main limitations that need to be addressed. First, they are typically designed for a specific algorithm with a predefined and limited set of parameters. Second, they often rely on the original versions of metaheuristic algorithms, which have been reported in the literature to be inefficient and prone to getting trapped in local optima \cite{lohvithee2021ant, rajwar2023exhaustive}.

\begin{figure*}[!t]
	\centering
		\includegraphics[width=1\linewidth]{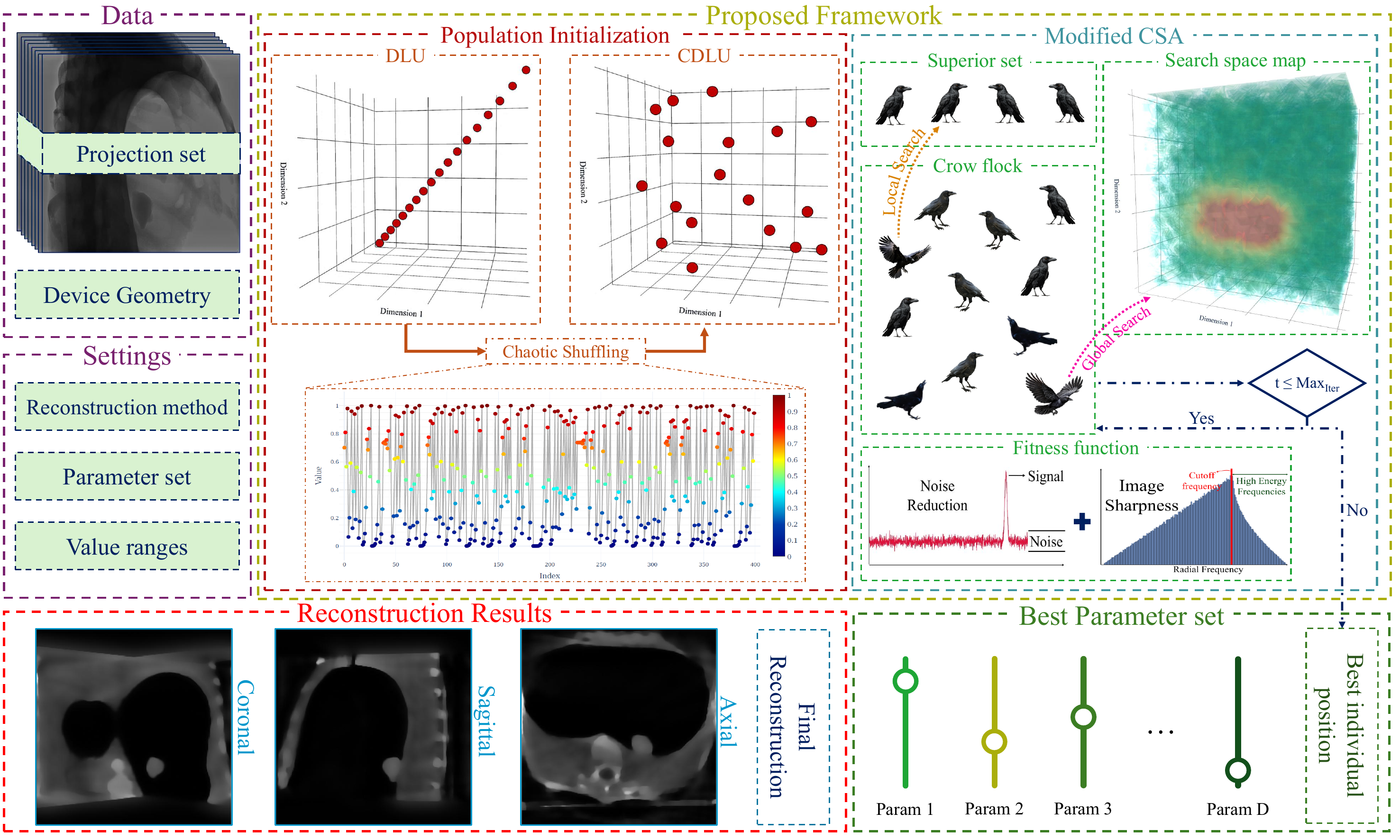}
	\caption{Overall automatic parameter optimization framework}
	\label{FIG:1}
\end{figure*}

In this paper, to overcome the limitations related to the lack of simultaneous tuning of multiple tunable parameters, the limitations of conventional metaheuristic optimization methods, and the need for reference data, an enhanced version of the crow search algorithm (CSA) \cite{askarzadeh2016novel}, referred to as the Search Space Aware CSA (SSA-CSA), with a more intelligent population initialization is proposed to tune the entire parameter set of CBCT iterative reconstruction algorithms. Figure \ref{FIG:1} illustrates the overall framework. To the best of our knowledge, this is the first general framework for multi-parameter optimization applicable to all iterative reconstruction algorithms on CBCT. Overall, the main contributions of this work are as follows:

\begin{enumerate}

    \item Design of an automatic parameter optimization framework that does not require reference data and can simultaneously optimize the entire parameter set, making it practical for real clinical settings.

    \item Development of an enhanced CSA algorithm with modified exploitation, exploration, and a more efficient balance between local and global search, to find the optimal solution in a huge search space, along with the introduction of a new population initialization method to generate the initial flock of crows more intelligently, addressing the limitations of original CSA getting stuck in local optima.
    
    \item Comprehensive evaluation of the proposed framework on three imaging machines and four real world datasets to demonstrate its generalizability across different scenarios and metrics.
    
\end{enumerate}

The rest of this paper is organized as follows. Section \ref{S:relatedworks} reviews related work. In Section \ref{S:Back}, some basic core concepts of the paper are overviewed. Section \ref{S:method} provides a comprehensive description of the proposed method. Section \ref{S:exp} presents the experimental setup and analysis of the results. Section \ref{S:diss} discusses the experimental findings, and Section \ref{S:conc} concludes the paper with suggestions for future research.

\section{Related Works}\label{S:relatedworks}

There are few works that perform automatic parameter optimization for iterative image reconstruction in CT and CBCT modalities. In this section, the most related and similar approaches to this study are reviewed. These works were selected based on the relevance of their optimization approach or the reconstruction method for which they optimized the hyperparameters.

In 2011, Xu et al. \cite{xu2011using} conducted one of the earliest attempts in this field. The main focus of their paper was to use graphical processing units (GPU) to accelerate iterative reconstruction methods. They then addressed the issue of tuning hyperparameters of the Ordered-Subsets Simultaneous Iterative Reconstruction Technique (OS-SIRT) \cite{xu2010efficiency} using a modification of the well-known genetic algorithm, NSGA-II \cite{deb2000fast}. However, the NSGA-II itself introduces additional tunable parameters that require investigation, and the objective function used in the study is entirely reference-based.

Zheng et al. \cite{zheng2012searching} attempted to find the optimal hyperparameters of OS-SIRT, but with the addition of smoothing and a non-local means (NLM) filter to balance dose, image quality, and reconstruction speed. Ant-Colony Optimization (ACO) \cite{dorigo2007ant} was used to identify the best set among other candidates. In an extended study by the same team \cite{zheng2013dqs}, a visual interface was added to help users understand and balance the trade-offs, along with a knowledge-based system to serve as an advisor to the human operator. However, the evaluations were relatively shallow, and the metrics used do not fully reflect human visual system assessment. This can lead to results that the framework considers optimal, but which may not be suitable from a clinician’s perspective.

In \cite{lohvithee2017parameter}, the authors proposed an adaptive version of projection-controlled steepest descent (PCSD) \cite{liu2015reconstruction}, referred to as AwPCSD. Also, they employed a semi brute-force approach to identify the optimal parameter set. This approach is described as semi brute-force because, although it resembles an exhaustive search, many candidate cases are pruned. Specifically, the algorithm starts with an arbitrary parameter set; then, all parameters except one are fixed, and a search is performed over the remaining free parameter. The best value for that parameter is selected and fixed, and this process is repeated for all parameters. Despite its contributions, the study has some limitations. First, the semi brute-force approach is computationally expensive and still risks overlooking the true optimal parameter set. Second, the evaluation relies on reference-based comparisons and non-standard metrics such as correlation coefficient (CC) and root mean square (RMS), making that not applicable for scenarios where reference data is not available.

With the recent advancements in deep learning and its integration with reinforcement learning, \cite{shen2018intelligent} introduced a deep reinforcement learning–based parameter optimization approach to determine the optimal regularization value for the alternating direction method of multipliers (ADMM) \cite{boyd2011distributed} algorithm. The core idea is to extract an image patch centered on each pixel in the reconstruction and feed it into a parameter tuning policy network (PTPN). Based on the current system state, the network selects one of five possible actions for adjusting the regularization parameter at that pixel: no change, or an increase/decrease of 10\% or 50\%. The restricted set of actions available to the PTPN constrains the optimization process. Also, the PTPN architecture is designed to tune only a single parameter, making it unsuitable for optimizing additional hyperparameters.

Lohvithee et al. \cite{lohvithee2021ant} proposed the most closely related approach to this paper. They utilized ACO to find the best parameter set for the AwPCSD algorithm. Overall, using this algorithm could find the parameters in a way that the image quality was better than arbitrary setting and cross-validation methods. However, this study has some major limitations. First, their optimization is based on a reference-based approach. Also, although AwPCSD needs six parameters to be set, in this work they fixed three of them and for the remaining three tunable parameters, have limited the number of choices, which makes the search space very shallow. Finally, the original ACO usually gets stuck in local optima \cite{dorigo2005ant}.

\section{Background}\label{S:Back}
This paper is based on two main concepts, namely iterative image reconstruction methods for CBCT for which we are trying to automate their hyperparameter tuning and CSA which is the foundation of our proposed bio-inspired metaheuristic and we aim to improve its performacnce. Understanding these two bases is necessary to fully comprehend the proposed framework and its objectives. 

\subsection{Overview of iterative image reconstruction}

In image reconstruction, the system is commonly modeled as a linear equation (Eq. \ref{Eq:1}). The goal is to recover an unknown image $x \in \mathbb{R}^{m}$, where $m$ is the total number of voxels in the image, from the observed data $b \in \mathbb{R}^{n}$ affected by additive noise $e \in \mathbb{R}^{n}$ where $n$ is the total number of measured pixels in the detector. The system matrix $A \in \mathbb{R}^{n \times m}$ describes how the image is transformed into the measurements.

\begin{equation}
\centering
\label{Eq:1}
Ax + e = b
\end{equation}

In simple linear algebra terms, $x$ could ideally be computed by directly inverting $A$. However, this is not feasible in practice because $A$ is ill-conditioned in CBCT. In such cases, $A$ has a rank lower than its dimension and thus even small perturbations in $b$ can result in large errors in $x$. 

Iterative reconstruction methods aim to approximate $x$ by minimizing a cost function that balances data fidelity with prior constraints, also known as regularization. The general form is given in Eq. \ref{Eq:2}.

\begin{equation}
\centering
\label{Eq:2}
\hat{x} = \underset{x}{argmin}~\mathrm{|| Ax - b||}_{2}^{2} + \Gamma R(x),
\end{equation}

where, $R(x)$ denotes the regularization term, while $\Gamma$ is the regularization parameter that controls the trade-off between data fidelity and regularization. Often for cases of undersampled or noisy data, the total variation (TV) norm of the image is used as a regulariztaion, $R(x) = ||x||_{TV}$.

ASD-POCS \cite{sidky2008image} is a well-known iterative algorithm that solves equation \ref{Eq:2} for convex sets, like the ones that arise with the TV norm. The solution is posed as constrained minimization as expressed in Eq. \ref{Eq:3}, while satisfying the two constraints shown in Eq. \ref{Eq:4}.

\begin{equation}
\centering
\label{Eq:3}
x^* = argmin~||x||_{TV}
\end{equation}

\vspace{-5mm}

\begin{equation}
\centering
\label{Eq:4}
||Ax - b|| \leqslant \epsilon~and~x\geqslant 0
\end{equation}

where $\epsilon$ represents the maximum allowed error between the predicted and observed projection data. The second constraint ensures that voxel intensities remain non-negative. In each iteration, ASD-POCS estimates $x^*$ and computes its predicted projections via forward projection. It then updates $x^*$ to make these predicted projections as close as possible to the actual measured projections. In the second step, it improves image quality by enforcing smoothness and preserving edges to reduce noise and artifacts.

The two-stage approach of ASD-POCS is employed by PCSD \cite{liu2015reconstruction}. However, the step size in the steepest descent process is adjusted adaptively, reducing the number of tunable hyperparameters in ASD-POCS. Additionally, AwPCSD \cite{lohvithee2017parameter} is a modified version of PCSD, where the TV norm is replaced by an adaptive-weighted TV norm to achieve better edge preservation.

Another powerful regularization term for highly undersampled projection data is prior image constrained compressed sensing (PICCS) \cite{chen2008prior}, which can be solved using the ASD-POCS algorithm and requires the same hyperparameters. However, the optimization formulation of PICCS (Eq. \ref{Eq:P1}) incorporates prior knowledge $x_p \in \mathbb{R}^{M}$. Where $\Psi_1$ and $\psi_2$ denote sparsifying transforms, while $\rho$ controls the relative contribution of the prior-based regularization term. The use case of PICCS differs from other iterative algorithms, as it is applied when relatively accurate prior knowledge of the object is available, and this prior information helps compensate for the missing data caused by sparse projection angles. Many studies have demonstrated its effectiveness when applicable \cite{hatamikia2024source, chen2011time, hastings2025real}.

\begin{equation}
\label{Eq:P1}
\begin{aligned}
\hat{x} = \underset{x}{\arg\min} \; & \left\{ 
\rho \|\Psi_1 (x - x_p)\|_{\ell_1} 
+ (1-\rho) \|\Psi_2 x\|_{\ell_1} 
\right\} \\
& \text{s.t.} \quad Ax = b
\end{aligned}
\end{equation}

ASD-POCS, AwPCSD, and PICCS produce high-quality reconstructions even with very limited projection data. However, these algorithms are highly dependent on their hyperparameters. Table \ref{Tab:1} presents the required hyperparameters for both algorithms, along with a short description and their typical value ranges. There is a huge search space of possible parameter settings for all mentioned algorithms, making manual tuning impractical.

\begin{table}[!t]
\centering
\caption{Tuneable Hyperparameters of ASD-POCS, AwPCSD, and PICCS}
\label{Tab:1}
\resizebox{0.8\columnwidth}{!}{%
\begin{tabular}{c|c|c|c}
\hline
\textbf{$Parameter$} & \textbf{$Algorithm$}                                       & \textbf{$Description$}                                         & \textbf{\begin{tabular}[c]{@{}c@{}}$Value range$\\ {[}$min, max, step${]}\end{tabular}} \\ \hline
\textbf{$Max\_iter$} & \begin{tabular}[c]{@{}c@{}}ASD-POCS\\AwPCSD\\PICCS\end{tabular} & Maximum algorithm iterations                                 & \begin{tabular}[c]{@{}c@{}}{[}5, 50, 1{]}\\ 46 values\end{tabular}                  \\ \hline
\textbf{$TV\_iter$}  & \begin{tabular}[c]{@{}c@{}}ASD-POCS\\ AwPCSD\\PICCS\end{tabular} & \begin{tabular}[c]{@{}c@{}}Number of TV minimization \\ steps per iteration\end{tabular}                & \begin{tabular}[c]{@{}c@{}}{[}5, 50, 1{]}\\ 46 values\end{tabular}                  \\ \hline
\textbf{$\epsilon$}         & \begin{tabular}[c]{@{}c@{}}ASD-POCS\\AwPCSD\\PICCS\end{tabular} & \begin{tabular}[c]{@{}c@{}}Data inconsistency tolerance\\ parameter\end{tabular}                       & \begin{tabular}[c]{@{}c@{}}{[}50, 1500, 10{]}\\ 146 values\end{tabular}             \\ \hline
\textbf{$\alpha$}         & \begin{tabular}[c]{@{}c@{}}ASD-POCS\\PICCS\end{tabular}                                                  & \begin{tabular}[c]{@{}c@{}}Initial steepest descent step\\ size\end{tabular}                          & \begin{tabular}[c]{@{}c@{}}{[}0.0001, 0.1, 0.0001{]}\\ 1000 values\end{tabular}     \\ \hline
\textbf{$\alpha\_red$}    & \begin{tabular}[c]{@{}c@{}}ASD-POCS\\PICCS\end{tabular}                                                  & \begin{tabular}[c]{@{}c@{}}Reduction factor of steepest\\ descent step size\end{tabular}               & \begin{tabular}[c]{@{}c@{}}{[}0.9, 0.99, 0.01{]}\\ 10 values\end{tabular}           \\ \hline
\textbf{$\lambda$}         & \begin{tabular}[c]{@{}c@{}}ASD-POCS\\ AwPCSD\\PICCS\end{tabular} & \begin{tabular}[c]{@{}c@{}}Sets the value of the hyperparameter\\ for the SART iterations\end{tabular}  & \begin{tabular}[c]{@{}c@{}}{[}0.9, 0.99, 0.01{]}\\ 10 values\end{tabular}           \\ \hline
\textbf{$\lambda\_red$}    & \begin{tabular}[c]{@{}c@{}}ASD-POCS\\ AwPCSD\\PICCS\end{tabular} & \begin{tabular}[c]{@{}c@{}}Reduction factor of SART\\ hyperparameter\end{tabular}                      & \begin{tabular}[c]{@{}c@{}}{[}0.9, 0.99, 0.01{]}\\ 10 values\end{tabular}           \\ \hline
\textbf{$r\_max$}    & \begin{tabular}[c]{@{}c@{}}ASD-POCS\\PICCS\end{tabular}                                                  & \begin{tabular}[c]{@{}c@{}}The maximum ratio of change\\ by TV minimization\end{tabular}              & \begin{tabular}[c]{@{}c@{}}{[}0.9, 0.99, 0.01{]}\\ 10 values\end{tabular}           \\ \hline
\textbf{$\delta$}         & AwPCSD                                                   & \begin{tabular}[c]{@{}c@{}}Scale factor for adaptive-weighted\\ TV norm\end{tabular}                   & \begin{tabular}[c]{@{}c@{}}{[}0.005, 2, 0.005{]}\\ 400 values\end{tabular}          \\ \hline
\end{tabular}
}
\end{table}

\subsection{Overview of crow search algorithm}

CSA \cite{askarzadeh2016novel} is a bio-inspired metaheuristic algorithm that simulates natural processes or animal instincts to solve complex problems. CSA and its variants have shown strong performance in high-dimensional engineering tasks, including feature selection \cite{samieiyan2022solving, anter2020feature} and medical image segmentation \cite{lenin2020fuzzy}.

Crows live in flocks, hiding their food while trying to follow other crows to locate and steal from their hiding places. This social behavior forms the basis of the CSA. The algorithm models both local and global search through two distinct scenarios. In the local search scenario, crow $i$ follows crow $j$ and moves closer to crow $j$'s hiding place, as formulated in Eq. \ref{Eq:5}. In the global search scenario, if crow $j$ finds out that it is being followed, it flies away. Consequently, crow $i$ moves randomly in search of another hiding place.

\begin{equation}
\centering
\label{Eq:5}
X_i^{t+1} = X_i^{t} + r_i \cdot L \cdot (M_j^{t} - X_i^{t}) 
\end{equation}

In Eq. \ref{Eq:5}, $X_i^{t+1}$ represents the position of crow $i$ after following hiding place of crow $j$, $r_i^t \in [0, 1]$ is a uniformly distributed random number, $L$ denotes the "flight length" or step size of the algorithm, and $M_{j}^{t}$ is the memory, or hiding place, of crow $j$ at iteration $t$ and it is updated at each iteration if the crow's new position is better than its current memory.

In this paper, the task of finding the optimal parameter set for an iterative reconstruction method is formulated into CSA. Specifically, the position of each crow is represented by a $1\times D$ vector, where each element corresponds to the value of a single hyperparameter, and $D$ is the total number of hyperparameters to be optimized.

\section{Materials and Methods} \label{S:method}
In this section, we detail the proposed framework for parameter optimization for CBCT iterative reconstruction methods by explaining the novel advanced population initialization, the fitness formulation, and the modifications to the original CSA, including local and global search strategies and the balancing method between them. We also provide a detailed explanation of the imaging systems and real datasets used to evaluate the proposed SSA-CSA.

\subsection{Proposed population initialization}

Population initialization is a critical step in population-based algorithms \cite{agushaka2022initialisation}. An effective initialization strategy is needed to ensure search space coverage, promote solution diversity, and accelerate convergence. In most original bio-inspired metaheuristics, initialization is performed randomly. However, random initialization has two major drawbacks. First, in purely random initializations drawn from a uniform distribution, it is not guaranteed that the initialization covers the space evenly. Second, since each run begins with a completely new population, the algorithm’s results require additional investigation to account for variability.

Several methods for more effective population initialization have been proposed \cite{tharwat2021population}. Nevertheless, when the search space is huge and the population size must remain limited for computational efficiency, two approaches stand out. These methods ensure that candidates can appropriately represent the entire search space.

Latin Hypercube Sampling (LHS) \cite{mckay2000comparison} is a widely used stratified sampling method for covering high-dimensional search spaces. The idea is to divide each dimension of the space into equally sized intervals, and then sample each interval once. In this way, LHS guarantees uniform coverage in the marginal distributions of all dimensions. However, LHS does not ensure that the entire multi-dimensional space is uniformly covered; in high-dimensional problems with a limited number of samples, it is still possible that some regions of the space remain under-sampled while others are relatively dense. Diagonal Linear Uniform (DLU) initialization \cite{li2021improved} can be considered a particular case of LHS. In DLU, rather than randomly permuting the intervals, samples are selected along the diagonal. With DLU, it is guaranteed that the entire search space is covered more evenly.

However, the output of DLU has a characteristic that makes it less feasible for the target task of this paper, in which the samples are ordered, meaning that the first sample has the lowest values in all dimensions, and the last sample has the highest values in all dimensions. Overall, some hyperparameters of iterative reconstruction should be near their low values, while others should be near their maximum values, and DLU does not provide this combination, which slows down the convergence of the algorithm. To address this problem, in this paper, after generating the population using DLU, the values of each dimension will be reordered across all individuals.

The shuffling technique used in this paper is not random, due to the mentioned drawbacks of using randomness in initialization. Instead, a sine chaos map is employed, which is a deterministic sequence of numbers that behaves like random. The formulation of the sine map sequence is given in Eq. \ref{Eq:6}, where $C^t$ denotes the value of the sine map sequence for each
 iteration $t$.

\begin{equation}
\label{Eq:6}
\begin{split}
& C^0= 0.7, \\
& C^t = sin(\pi \cdot C^{t-1}),
\end{split}
\end{equation}

Algorithm \ref{alg:1} presents the proposed Chaotic DLU (CDLU) for population initialization. With CDLU, we can ensure that the search space is evenly covered, and the problem of having ordered individuals is addressed.

\begin{algorithm}
\caption{Chaotic Diagonal Linear Uniform Initialization}
\label{alg:1}
\begin{algorithmic}[1]
\REQUIRE $N=$ Population size ,\\ $D=$ Dimension size ,\\  $[min_d, max_d]=$ Range for each dimension
\STATE $C = [C^0, C^1, ..., C^{N\cdot D}, \text{where } C^t \text{ is eq. \ref{Eq:6}}$
\STATE $\text{Chaotic\_indices} \leftarrow \lceil C \cdot N \rceil$ \COMMENT{Scale values}
\FOR{$d = 1$ to $D$}
    \STATE $step \leftarrow \dfrac{max_d - min_d}{N - 1}$ \COMMENT{Interval size}
    \STATE $X^{\text{0}}(d) \leftarrow \text{Arithmetic-progression($min_d$, $max_d$, }step)$
    
    \STATE $\text{Reorder $X^{0}(d)$ based on Chaotic\_indices}$ \COMMENT{Apply chaotic shuffling}
\ENDFOR
\RETURN $X^{0}$
\end{algorithmic}
\end{algorithm}

\begin{figure}[]
	\centering
		\includegraphics[width=0.6\textwidth]{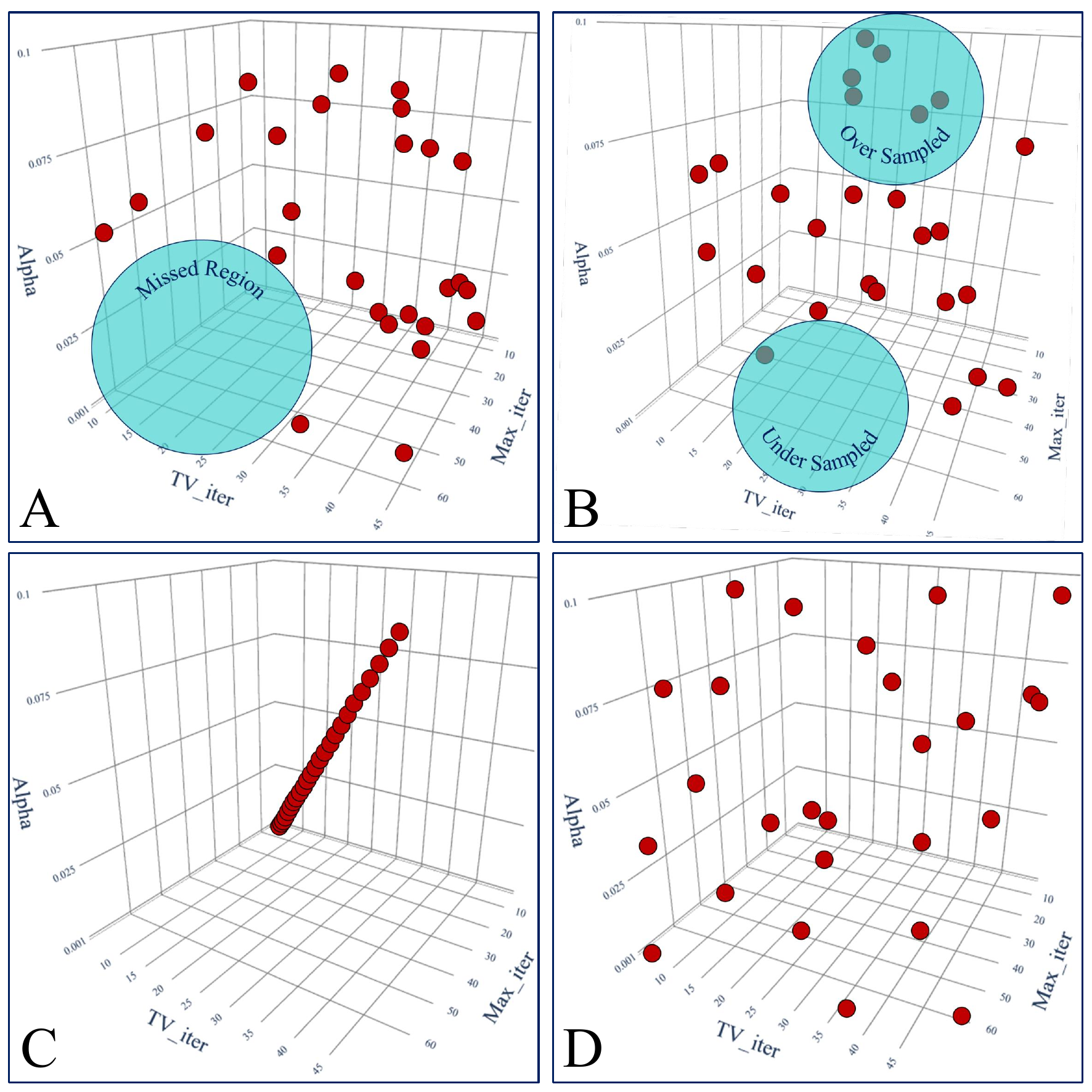}
	\caption{(A) Random Initialization, (B) LHS initialization, (C) DLU initialization, and (D) Proposed chaotic DLU initialization}
	\label{FIG:2}
\end{figure}

The efficiency of CDLU is examined in the results section by comparing the solutions obtained using the same search algorithm but with different initialization methods. To provide an overall view of how these methods explore the search space, Figure \ref{FIG:2} illustrates the distribution of 25 individuals using $random$, $LHS$, $DLU$, and $CDLU$ initializations. For visualization purposes, only the three most important parameters, $alpha~(\alpha)$, $max\_iter$, and $TV\_iter$, are presented.

\subsection{Proposed fitness function}

The most important part of an optimization process is the evaluation of different solutions, which drives the progress of optimization. Defining an inappropriate objective function, also known as the fitness function, can lead the optimization algorithm to converge to a solution that is not suitable for the problem. In some cases, defining the objective function is quite straightforward; For example, minimizing a benchmark function with a known formulation, where different solutions can be evaluated directly by computing the function’s output, which then serves as the objective value.

However, defining an objective function for optimizing the parameter set of an iterative reconstruction algorithm is more challenging for two main reasons. First, the objective value cannot be derived directly from the candidate parameter set; instead, it must be evaluated through its application within the reconstruction algorithm. Second, the output of the reconstruction algorithm is not a simple scalar value, but rather a volumetric image which needs an image quality assessment metric which represent a numerical evaluation of the output. 

The Signal-to-Noise Ratio (SNR) is one of the most common metrics for assessing the noise level of an image and can be calculated using Eq. \ref{Eq:7}. In this paper, we also aim to improve the SNR of the output images. However, a limitation of SNR is that it does not account for image sharpness which is an important characteristic of medical images. If the fitness function relies solely on SNR, the proposed method may converge to a parameter set that produces overly smooth and blurry reconstructed images, with flat texture and no visible edges.

\begin{equation}
\centering
\label{Eq:7}
\mbox{SNR}(I) = \frac{1}{K} \sum_{k=1}^{K}\frac{\mu(I_k)} {\sigma(I_k)}
\end{equation}

In Eq. \ref{Eq:7}, $I$ denotes the 3D volume reconstructed using the candidate parameter set $X$ ($I = Recon(b, X)$) and $K$ represents the total number of slices within the volume. $I_k$ refers to the $k$-th 2D slice, while $\mu(I_k)$ and $\sigma(I_k)$ denote the mean and standard deviation of the $k$-th slice, respectively.

To achieve a better representation of reconstruction output, we design a multi-objective fitness function that balances noise reduction, represented by SNR, and image sharpness. Various approaches exist for evaluating sharpness, but assessing the high-frequency energy ratio (HFER) over the entire volume provides a reliable measure. Let $\mathcal{F}\{I_k\}(u, v)$ be the discrete Fourier transform of one slice, where $u$ and $v$ represent spatial frequencies. The Power Spectrum $P\{I_k\}(u, v)$ is defined as in Eq. \ref{Eq:F1}.

\begin{equation}
\centering
\label{Eq:F1}
P\{I_k\}(u, v) = \left| \mathcal{F}\{I_k\}(u, v) \right|^2
\end{equation}

On the other hand, the radial frequency coordinate $Q\{I_k\}(u, v)$ from the center of the spectrum can be computed using Eq. \ref{Eq:F2}.

\begin{equation}
\centering
\label{Eq:F2}
Q\{I_k\}(u,v) = \sqrt{u^{2} + v^{2}}
\end{equation}

If we consider $Q_{max}$ as the maximum possible frequency radius in the grid, the cutoff frequency $f_c$ can be obtained using Eq. \ref{Eq:F3} based on defined cutoff ratio $\gamma$.

\begin{equation}
\centering
\label{Eq:F3}
f_{c} = \gamma \cdot Q_{max}
\end{equation}

The total energy ($E_{total}$) is defined as the sum of the power spectrum over the entire grid, whereas high-frequency energy ($E_{high}$) is computed as the sum of the power spectrum only where the radial frequency exceeds the cutoff frequency. Therefore, HFER can be calculated as the average ratio of high-frequency energy to total energy across all $K$ slices of the reconstructed volume (Eq. \ref{Eq:F6}).

\begin{equation}
\centering
\label{Eq:F6}
\mbox{HFER}(I) = \frac{1}{K}\sum_{k=1}^{K}\frac{\sum_{u,v}P\{I_k\}(u, v) \cdot \chi(Q\{I_k\}(u,v) > f_{c})}{\sum_{u,v}P\{I_k\}(u, v)}
\end{equation}

In Eq. \ref{Eq:F6}, $\chi(\cdot)$ is a binary function that equals one if the condition is true and zero otherwise.

Accordingly, the overall fitness function in this work is formulated as a weighted summation of SNR and HFER (Eq. \ref{Eq:8}). Since the optimization is defined as a minimization problem, the fitness function is expressed in minimization form.  The parameters $\eta$ and $\xi$ can be easily tuned depending on the desired outcome. A higher value of $\eta$ leads to smoother images with reduced noise, whereas a higher ratio of $\xi$ results in sharper images at the cost of increased noise.


\begin{equation}
\label{Eq:8}
\mbox{Fitness}(I) 
= (\eta \cdot \frac{1}{\mbox{SNR}(I)}) + (\xi \cdot (1 - \mbox{HFER}(I))
\end{equation}

In this work, the design of the fitness function emphasizes independence from any reference image. This allows the optimization to be performed on data without available reference to make the proposed method more practical. In real clinical cases, reference reconstructions are rarely accessible for parameter tuning using reference-based metrics. In addition, the proposed fitness function is computationally fast and more explainable than complex image quality metrics, such as learning-based image quality assessments \cite{lee2025low}, which are computationally expensive and difficult to interpret. A correlation assessment between the proposed fitness and other learning-based image quality assessments has been performed in Appendix \hyperref[App_A]{A}.

\subsection{Improved local search strategy}

In the CSA, some crows are selected to perform a local search. Local search, also known as exploitation, is the process in which individuals in the population explore the search space based on their surroundings or neighbors. The efficiency of local search depends primarily on two factors: the movement mechanism and the neighborhood definition. The movement mechanism of CSA in local search is described in Eq. \ref{Eq:5}, which incorporates a randomness component. Studies have shown that using chaotic maps in this step can significantly improve the convergence speed and overall performance of CSA \cite{ouadfel2020enhanced, sayed2019feature}.

On the other hand, the neighborhood concept in CSA is defined randomly. Specifically, when a crow performs local search, it randomly selects another crow to follow. However, this approach has a major drawback: in some cases, a crow with poor performance may be selected as the neighbor. Consequently, the following crow is misled toward a poor position, and if this issue is not properly addressed, the overall results may progressively deteriorate.

In this paper, inspired by previous works, the random element is replaced with the Sine chaotic map (eq. \ref{Eq:6}). In addition, we propose a more effective neighborhood concept. At the beginning of each iteration, a superior set $S$ of crows is selected, as outlined in Algorithm \ref{alg:2}. First, the set is filled using the Pareto front technique, which is more effective than other multi-objective combination methods because it considers each objective separately and identifies the crows that are not dominated by any other crow with respect to their corresponding $objective~vector$. However, if the number of crows in the Pareto front is not sufficient to form the superior set, the remaining slots are filled based on the sorted fitness values of the crows. 

Once this superior set is formed, when a crow needs to select another crow to follow, it chooses one from the superior set. This approach mitigates the risk of following a poorly performing crow. The novel movement mechanism of crows is defined as follows:

\begin{equation}
\label{Eq:9}
\centering
X_i^{t+1} = X_i^{t} + C_i^{t} \cdot L \cdot (S_j^{t} - X_i^{t}) 
\end{equation}

where $C_i^{t}$ denotes the chaos value of crow $i$, and $S_j^t$ represents the randomly selected crow $j$ from the superior set at iteration $t$. The number of slots in the superior set is determined based on $\kappa^t$, where $\kappa^t$ denotes the selection ratio at iteration $t$ and gradually decreases over the algorithm’s iterations by a constant rate $\kappa_{red}$, enhancing the quality of the crows in the set and thereby enabling more effective local search as the algorithm approaches its end.

\begin{algorithm}
\caption{Superior Set Selection at iteration $t$}
\label{alg:2}
\begin{algorithmic}[1]
\REQUIRE $\kappa^{t-1}=$ \text{selection\_rate at iteration $t-1$}\\ 
         $\{\text{Fitness}(Recon(X_i)), objective~vector_i \}$ for each $X_i$ (Eq. \ref{Eq:8}),\\
         $T$=Maximum iterations,\\
         $\text{Constant }\kappa_{red} = 1 - \left( \dfrac{1}{T} \right)$ \COMMENT{Reduction ratio}
\STATE $\kappa^t \leftarrow \kappa^{t-1} \cdot \kappa_{red}$ \COMMENT{update selection ratio}
\STATE $S \leftarrow \text{find\_pareto\_set}(objective~vector)$
\IF{$\text{size of $S$ did not reach }N \cdot \kappa^t$}
    \STATE $S \leftarrow \text{fill}(S, \text{sort(Fitness}))$
\ENDIF

\RETURN $S$
\end{algorithmic}
\end{algorithm}

\subsection{Novel search space aware global search}
Searching in a vast search space with a small population requires a more purposeful search strategy. In CSA, when a crow performs global search (exploration), it selects a completely random position in the search space. However, such randomness reduces the chance of following a meaningful strategy. Meanwhile, the overall concept of global search must be preserved to prevent the algorithm from getting stuck in local optima.

To address this, we add a feature called search space awareness to the crows. This means that when a crow explores, it has a general overview of the entire search space and is more likely to select a position that yields better solutions. To implement this feature, an $M$-dimensional weight map $\omega_t \in \mathbb{R}^{M}$ is initialized at the beginning of the algorithm. During the optimization, the weight map is updated using the positions of crows in the superior set by adding weight to their locations and to positions within $\pm$10\% of their values.

When a crow performs global search, it uses weighted random sampling instead of uniform random sampling. This ensures that regions of the search space with higher weights are more likely to be selected. Furthermore, the update factor $K_t$ of the weight map gradually increases by a factor of $\omega_{inc}$, placing greater emphasis on the regions around the most recent best-performing crows. 

\begin{figure}[]
	\centering
		\includegraphics[width=1\textwidth]{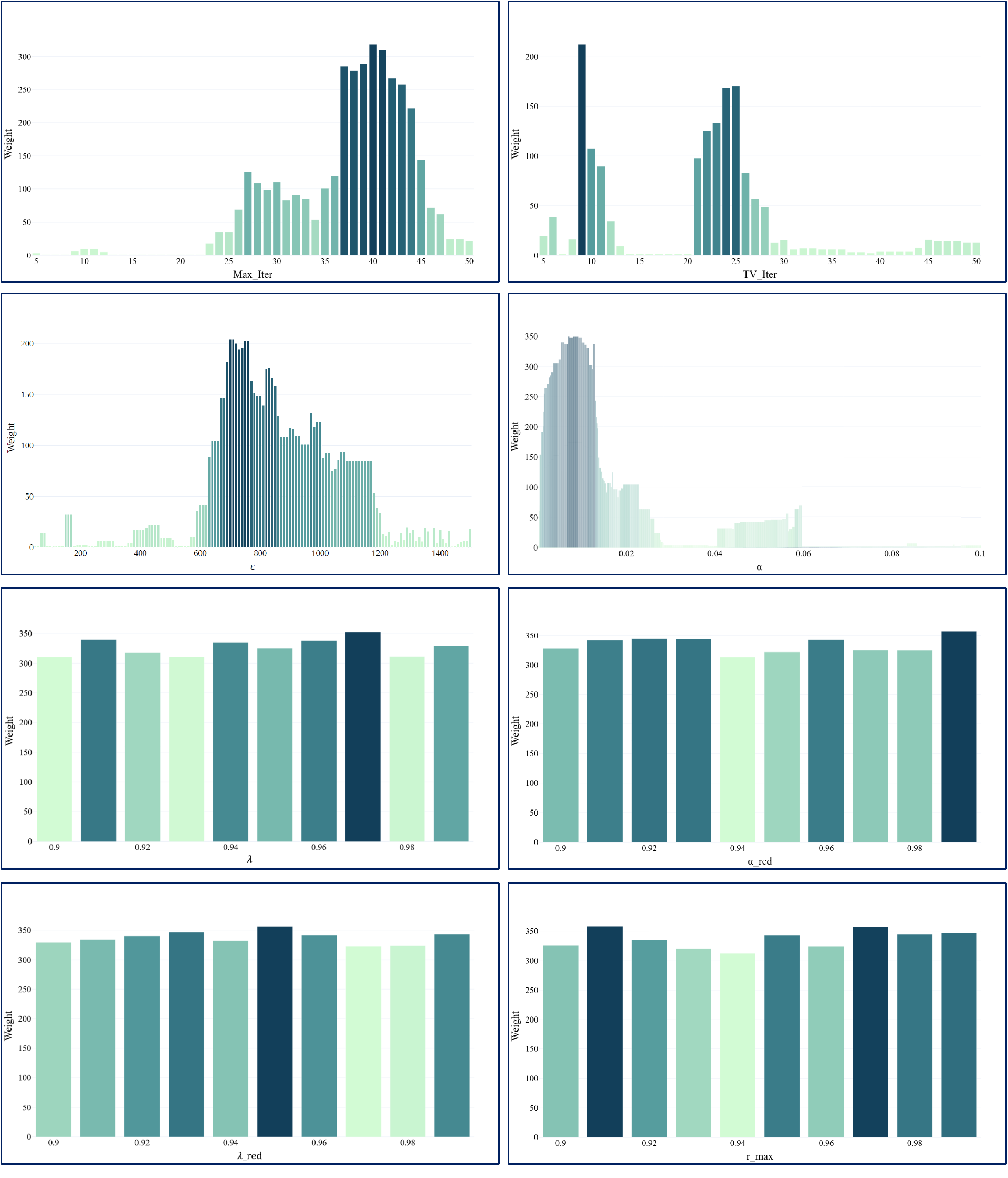}
	\caption{Weight map of search space after running the proposed SSA-CSA method to optimize ASD-POCS for a dataset. Each plot corresponds to one hyperparameter of the algorithm and indicates the weight of each value.}
	\label{FIG:3}
\end{figure}

Algorithm \ref{alg:3} presents the search space awareness process, and Figure \ref{FIG:3} illustrates an example of the final weight map of the entire search space for ASD-POCS after the algorithm completes. It can be observed that the weight map not only enables the proposed SSA-CSA to perform a more effective global search, but also provides the user with an overview of the different hyperparameters of the algorithm and their relative importance. For instance, in the example provided in Figure \ref{FIG:3}, it can be inferred which ranges of values for $\alpha$, $TV\_iter$, and $max\_iter$ yield better performance. For some other parameters, such as $\alpha_{red}$ and $\lambda_{red}$, there is a more uniform distribution in the weight map, meaning that they have less impact on the overall performance.

\begin{algorithm}[b]
\caption{Search space awareness technique at iteration $t$}
\label{alg:3}
\begin{algorithmic}[1]
\REQUIRE $\omega_{t-1}=\text{Weight map at iteration } t-1$,\\ $K_{t-1}=\text{Update value at iteration } t-1$,\\ $\omega_{inc}=\text{Increase rate } $
\STATE $K_t \leftarrow K_{t-1} \cdot \omega_{inc}$ \COMMENT{Increase update value}
\STATE $\omega_t \leftarrow \text{update}(S, K_t)$
\STATE $X_i^t \leftarrow \text{weighted sampling based on }\omega_t$ \COMMENT{Perform global search for crow$_i$}
\end{algorithmic}
\end{algorithm}

\subsection{Enhanced exploitation and exploration balance}

In all metaheuristic search algorithms, there must always be a balance between local search and global search. Excessive local search can cause the algorithm to get stuck in local optima, while excessive global search can prevent convergence. In CSA, this balance is achieved by generating a random number and comparing it with a fixed parameter known as awareness propability ($AP$). Although this approach introduces some balance, it is suboptimal because the condition is determined purely by chance. Consequently, a high-quality crow may fail to satisfy the condition and unnecessarily perform exploration, losing its position, whereas a low-quality crow may satisfy the condition and perform local search.

In this paper, inspired by our previous work \cite{samieiyan2022novel}, we modify this condition so that high-quality crows perform local search while low-quality crows perform global search. To achieve this, we incorporate the fitness value of crows into the decision-making process. In the proposed balancing technique, each crow's fitness is compared with that of the rest of the population. If its fitness is lower than a specified $AP_t$ percentage of the population at iteration t denoted by $fitness~threshold$, it performs local search; otherwise, it performs global search. In addition, the specified $AP$ percentage is gradually increased over time using an $AP_{inc}$ factor, which encourages the population to perform more local search than global search as the algorithm approaches the end. Algorithm 4 details the steps for deciding between local and global search.

\begin{algorithm}
\caption{Adaptive Local and Global Search balance}
\label{alg:4}
\begin{algorithmic}[1]
\REQUIRE $T=$ Maximum iterations 
\STATE $X^0 \leftarrow \text{CDLU population initialization}$\COMMENT{(Alg. \ref{alg:1})}
\STATE $M^0 \leftarrow X^0$\COMMENT{Memory initialization}
\STATE Initialize $AP^0$
\FOR{$i = 1$ to $N$}
\STATE $ Fitness_i^{0} \leftarrow \text{Fitness} (Recon(M_i^{0}))$\COMMENT{(Eq. \ref{Eq:8})}
\ENDFOR
\FOR{$t = 1$ to $T$}
\STATE $AP^t \leftarrow AP^{t-1} \cdot AP_{inc}$
    \FOR{$i = 1$ to $N$}
        \STATE $\text{fitness\_threshold} \leftarrow AP^t\text{-quantile of }Fitness^{t-1}$
        \IF{$\text{Fitness}_i^{t-1} < \text{fitness\_threshold}$}
            \STATE \COMMENT{Local Search (Eq. \ref{Eq:9})}
        \ELSE
            \STATE \COMMENT{Global Search (Alg. \ref{alg:3})}
        \ENDIF
        \STATE $ Fitness_i^{t} \leftarrow \text{Fitness} (Recon(X_i^{t}))$\COMMENT{(Eq. \ref{Eq:8})}
        \IF{$\text{Fitness}_i^{t} < \text{Fitness}_i^{t-1}$}
            \STATE \text{Update memory to new position}
        \ENDIF
    \ENDFOR
\ENDFOR
\end{algorithmic}
\end{algorithm}

\subsection{Datasets}
To ensure a comprehensive evaluation and an in-depth discussion of the results, four real-world datasets acquired by different imaging systems including Nikon Custom Bay X-ray CT, ImagingRing m (IRm), and Philips Allura to provide diverse clinical scenarios are employed in this study to demonstrate the effectiveness of the proposed SSA-CSA method. 

\subsubsection{SophiaBeads}
SophiaBeads \cite{coban2015sophiabeads} is a microCT scan of a plastic tube filled with Soda-Lime glass. This real dataset was acquired specifically to evaluate different reconstruction algorithms. The imaging was performed using a Nikon Custom Bay X-ray CT system. The dataset provides very high-resolution projections with a size of 1564$\times$1564. In the main dataset, the overall number of projections is 2048; however, in this work and the similar study \cite{lohvithee2017parameter}, 64 projections are used for iterative reconstruction methods. SophiaBeads is the only public dataset used in this study. For a deeper investigation, we used the simultaneous algebraic reconstruction technique (SART) \cite{andersen1984simultaneous} applied to the full projection data as the gold standard for this dataset.

\subsubsection{LinePairs CatPhan phantom}
The second dataset is an in-house dataset used to evaluate the results. The imaging machine for this dataset is a CBCT device called IRm, developed by medPhoton GmbH, which is specifically designed for image-guided radiotherapy and incorporates unique features such as non-isocentric imaging and a selectable region of interest \cite{keuschnigg2017nine}. IRm is equipped with a state-of-the-art flat-panel detector capable of acquiring high-resolution projections of 1440$\times$1440 pixels (2$\times$2  binning mode), with a pixel size of 0.3$\times$0.3 mm. A linepairs CatPhan phantom was scanned using IRm with 239 projections to simulate a low-dose acquisition scenario.

\subsubsection{Thorax phantom}
Hatamikia et al. \cite{hatamikia2020additively} developed a patient-specific anthropomorphic thorax phantom based on anonymized CT data, which was segmented and 3D-printed using PolyJet technology. The skeletal structures were filled with custom radiopaque mixtures of bone meal, epoxy, and polypropylene to replicate realistic Hounsfield Units for different body organs. In addition, in \cite{hatamikia2024source}, two 3D-printed objects were inserted into the phantom to serve as tumors and to evaluate reconstruction quality based on the visibility of these targets. The number of projections used in this study was 142, acquired with equal angular spacing using a Philips Allura FD20 Xper C-arm CBCT system.

\subsubsection{Brain phantom}\label{Brain_data_Sec}
The last dataset used in this study is a pair of two acquisitions of a brain phantom, performed without and with a metal screw inserted into the phantom. Each pair consists of a set of 473 projections and are taken by IRm machine. The aim of using this dataset is to validate the PICCS method \cite{chen2008prior}, which is a powerful algorithm for highly undersampled projection datasets. To apply PICCS, prior knowledge is required, which is usually obtained from projection set from the pre-operative stage of the object. In this case, the reconstruction from the full projection set without the metal screw, obtained using the optimized ASD-POCS method, is considered as the prior.

In this case, the post-operative data correspond to the projection set of the brain after screw insertion. To make this setting closer to a real clinical scenario with significant radiation dose reduction, only 20\% of the full projection set was sampled with equal angular spacing. This means that the PICCS algorithm is expected to reconstruct the brain with the screw clearly using only 95 projections, by exploiting the available prior knowledge.

\section{Experiments and results}\label{S:exp}
In this section, numerous studies and experiments are assessed to evaluate the entire proposed framework. The results section is mainly divided into two parts. In the first part, the focus is on one real dataset to conduct all ablation studies under different scenarios, compare the SSA-CSA with existing methods, and evaluate the effectiveness of each component of the framework. In the second part, additional real datasets are used to test and evaluate the generalizability and effectiveness of the proposed method on different real clinical data. The detailed solutions of all experiments conducted in this study are presented in Appendix \hyperref[App_B]{B}.

\subsection{Implementation and evaluation details}
\subsubsection{Implementation details}
The proposed framework was implemented using the Tomographic Iterative GPU-based Reconstruction (TIGRE) toolbox \cite{biguri2025tigre}, which provides most of the reconstruction algorithms for CT and CBCT. All code was written in MATLAB 2024 and executed on a computer with an NVIDIA RTX A4000 GPU and a 12-core AMD Ryzen processor. For all population-based metaheuristic experiments, the population size was set to 25 and the maximum number of iterations to 30. Furthermore, the values of $\eta$ and $\xi$ in Eq.~\ref{Eq:8} should be chosen based on the desired outcome of the framework and set experimentally to ensure a proper balance between the objective terms. In addition, in all of the following experiments, "manual setting" means that the algorithm was run with 10 to 15 different parameter sets in an attempt to improve the image quality, and the best result among these trials was reported. This simulates a real clinical scenario in which the operator cannot spend too much time tuning the parameters and instead selects the parameter set that performs relatively better than the others.

\subsubsection{Evaluation metrics}

Effective evaluation of medical images, in a way that the metric can accurately represent the quality level and maintain good agreement with human reader assessment, is a challenging task. Recently, learning-based approaches such as convolutional neural networks and transformers have been employed to predict a quality score corresponding to a medical image. In \cite{lee2025low}, several of these methods are introduced to assess the quality of low-dose CT images without requiring a reference. In this work, two of the best-performing models, namely CHILL@UK and RPI\_AXIS, are used to evaluate the results, along with classical evaluation metrics such as SNR, HFER, and fitness score, as introduced in the Methods section. The higher the values of the learning-based metrics are, the better the quality of the volume is. Since these metrics operate on single slices, we computed them as the average over all slices.

\subsection{Results on SophiaBeads dataset and ablation experiments}
As the proposed SSA-CSA incorporates several modifications to enhance the results such as introducing a multi-objective no-reference fitness function, a new non-random initialization, and a modified version of the original CSA, it is important to assess each modified component separately to demonstrate their effectiveness. To this end, a series of different scenarios are designed and evaluated in this section. For this part, all ablation studies focus on SophiaBeads reconstruction, as it is a publicly available dataset, allowing us to apply the evaluation metrics and interpret the results with respect to other related approaches.

\subsubsection{Analysis on fitness functions}\label{ss:fitness}
The fitness function plays a vital role in metaheuristic optimization and has a notable effect on the final solution found by the algorithm. In this experiment, we evaluate five different fitness functions to assess their impact on the final solution for the ASDPOCS hyperparameter set. The evaluated fitness functions are SNR, Laplacian variance, HFER, and the combinations of Laplacian variance and HFER with SNR. The Laplacian variance of a volume is calculated by applying a Laplacian filter and computing its variance. The sharper the image, the higher the variance will be.

\begin{figure}[]
	\centering
		\includegraphics[width=0.8\linewidth]{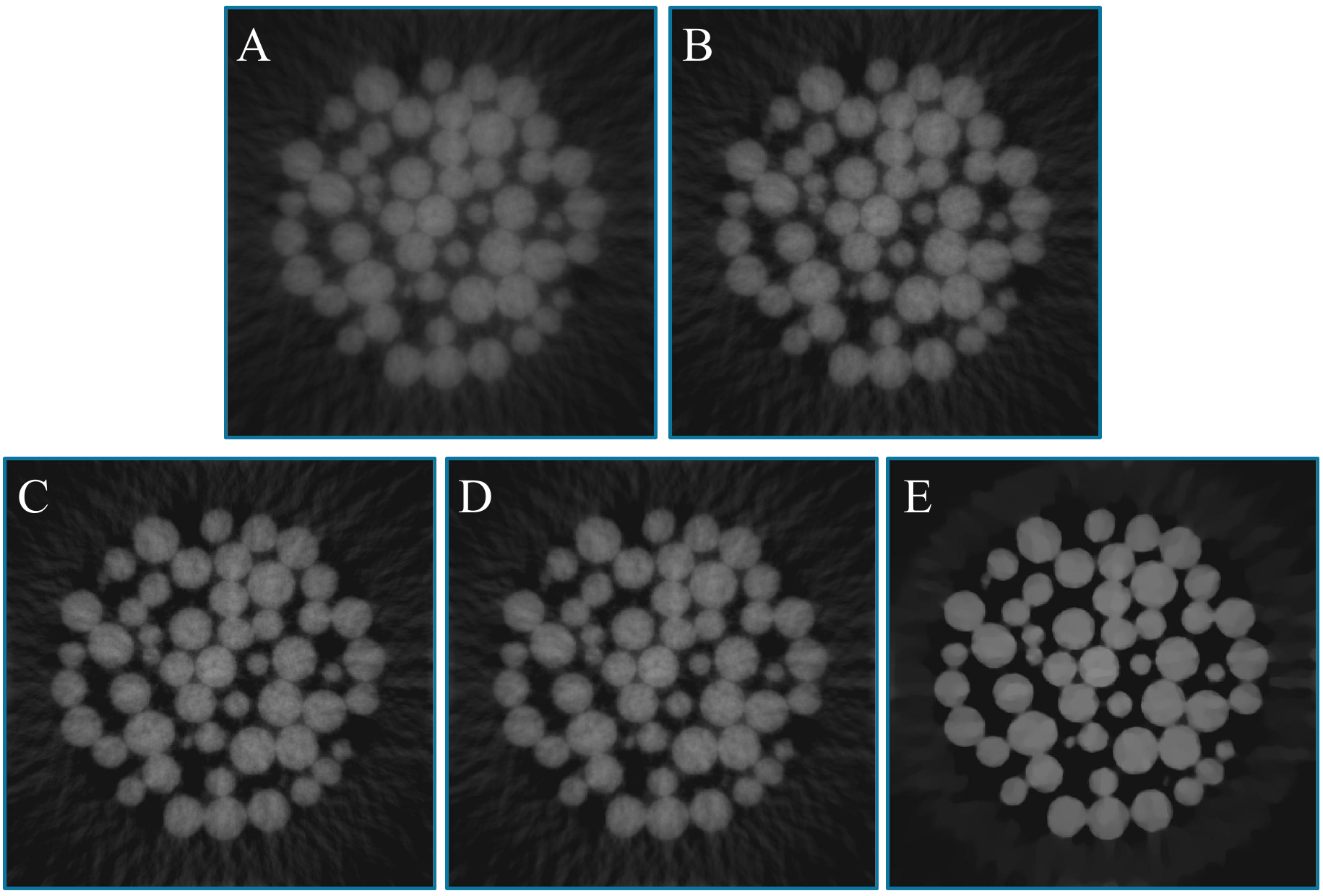}
	\caption{Qualitative results of (A) SNR, (B) Laplacian, (C) HFER, (D) SNR-Laplacian, and (E) proposed fitness functions}
	\label{FIG:4}
\end{figure}

Table \ref{Tab:2} presents the results of this experiment. As shown in classic metric evaluations, there is always a direct trade-off between SNR and HFER score, and with the same number of projections, it is impossible to improve both metrics simultaneously. Therefore, the goal of the optimization here is to find the best balance point. A notable difference can be observed in fitness and model-based metrics between the proposed fitness and the rest of the tested fitnesses, highlighting the effectiveness of the SNR-HFER fitness and its importance in the final solution. Figure \ref{FIG:4} illustrates the corresponding qualitative results of this experiment. In this figure, the SSA-CSA found a solution in which the edges of the circles are sharp, while there are no artifacts inside the circles, unlike the results obtained with HFER or Laplacian fitnesses.

\begin{table}[!t]
\centering
\caption{Performance of different fitness functions.}
\label{Tab:2}
\resizebox{0.8\columnwidth}{!}{%
\begin{tabular}{c|ccc|ccl}
\multicolumn{1}{l|}{\multirow{2}{*}{}}                                        & \multicolumn{3}{c|}{\textbf{Classic metrics}}                                          & \multicolumn{3}{c}{\textbf{Learning-Based Models}}                \\ \cline{2-7} 
\multicolumn{1}{l|}{}                                                         & SNR     & HFER & Fitness          & CHILL@UK                   & \multicolumn{2}{c}{RPI\_AXIS}        \\ \hline
\begin{tabular}[c]{@{}c@{}}SNR\\ Fitness\end{tabular}                         & 1.34170 & 0.15460                                                   & 4.32603          & 2.76648                    & \multicolumn{2}{c}{0.54072}          \\ \hline
\begin{tabular}[c]{@{}c@{}}Laplacian\\ Fitness\end{tabular}                   & 1.07590 & 0.23580                                                   & 4.08952          & 2.50858                    & \multicolumn{2}{c}{0.54818}          \\ \hline
\begin{tabular}[c]{@{}c@{}}HFER\\ Fitness\end{tabular}                   & 0.90320 & 0.30440                                                   & 3.90522          & 2.31846                    & \multicolumn{2}{c}{0.58378}          \\ \hline
\begin{tabular}[c]{@{}c@{}}SNR+Laplacian\\ Fitness\end{tabular}               & 0.94390 & 0.28710                                                   & 3.94965          & 2.79979                    & \multicolumn{2}{c}{0.54388}          \\ \hline
\begin{tabular}[c]{@{}c@{}}Proposed fitness\\      SNR+HFER\end{tabular} & 0.71330 & 0.37350                                                   & \textbf{3.80060} & \textbf{3.08410}           & \multicolumn{2}{c}{\textbf{0.73592}}
\end{tabular}
}
\end{table}

\subsubsection{Initialization methods comparison}\label{ss:init}
It is important to have an initial population that can cover the large search space of this study as uniformly as possible. In this experiment, the proposed CDLU initialization is compared with random, LHS, and DLU initialization. Another major objective of this experiment is to evaluate the stochastic nature of metaheuristics. To this end, each case was run 10 times to assess the mean and standard deviation (STD) of the different random initialization methods.

\begin{figure}[]
	\centering
		\includegraphics[width=0.65\textwidth]{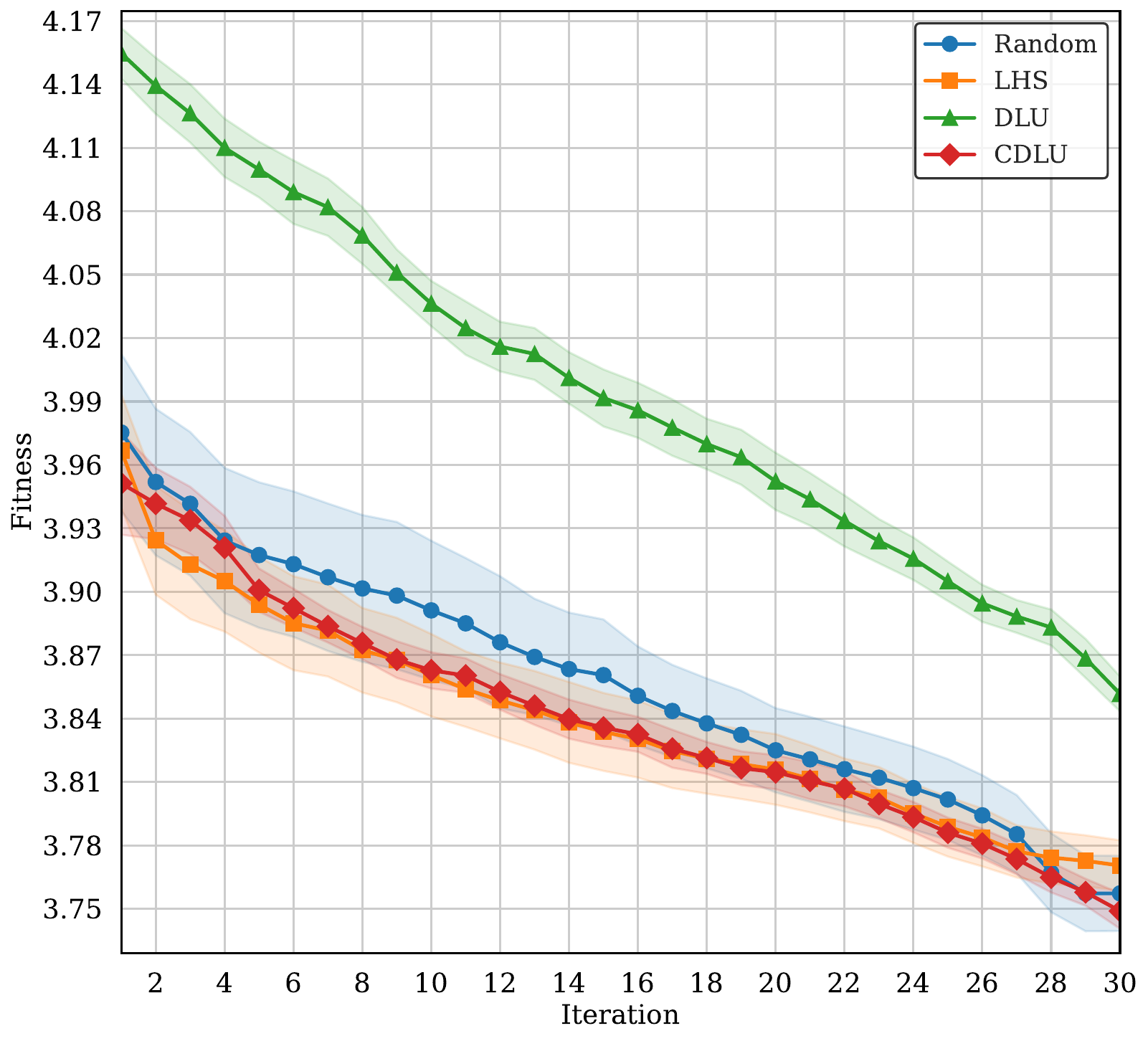}
	\caption{Fitness convergence using different initialization methods.}
	\label{FIG:5}
\end{figure}

According to the fitness convergence curves presented in \ref{FIG:5}, it can be clearly seen that, on average, DLU starts with a poor population due to the ordering of its values, which significantly affects its convergence speed. On the other hand, CDLU, LHS, and random initialization presented relatively similar convergence curves. However, as shown in Table \ref{Tab:3}, which reports the average of the best final solutions over 10 runs, it is evident that CDLU and random initialization outperform the other two methods, with very close average values. The main advantage of CDLU is its lower standard deviation (STD) compared to the random method. This indicates that CDLU achieves more stable convergence to the optimal solution.

The statistical evaluation was conducted only in this experiment because the stochastic effect in the proposed SSA-CSA is mainly due to the initialization step, as it employs more deterministic operations rather than random-based actions compared to the original CSA. In addition, CDLU is fully deterministic and produces the same population across all runs. In addition, from a computational perspective, running each case of all experiments multiple times would require excessive time.

\begin{table}[]
\centering
\caption{Performance of the proposed SSA-CSA with different initialization methods.}
\label{Tab:3}
\resizebox{0.8\columnwidth}{!}{%
\begin{tabular}{c|ccc|cc}
\multicolumn{1}{l|}{\multirow{2}{*}{}}                            & \multicolumn{3}{c|}{\textbf{Classic metrics}}                                          & \multicolumn{2}{l}{\textbf{Learning-Based Models}} \\ \cline{2-6} 
\multicolumn{1}{l|}{}                                             & SNR     & HFER & Fitness          & CHILL@UK                 & RPI\_AXIS               \\ \hline
\begin{tabular}[c]{@{}c@{}}Random\\ initialization\end{tabular}   &\begin{tabular}[c]{@{}c@{}}0.90571 \\ $\pm$ 0.0387
\end{tabular} & \begin{tabular}[c]{@{}c@{}}0.36678 \\ $\pm$ 0.0069\end{tabular}                                                   & \begin{tabular}[c]{@{}c@{}}3.62238 \\ $\pm$ 0.0178\end{tabular}          & \begin{tabular}[c]{@{}c@{}}2.94742\\ $\pm$ 0.1459\end{tabular}                  & \begin{tabular}[c]{@{}c@{}}0.74609\\ $\pm$ 0.0451\end{tabular}                 \\ \hline
\begin{tabular}[c]{@{}c@{}}LHS\\ initialization\end{tabular}      & \begin{tabular}[c]{@{}c@{}}0.86339 \\ $\pm$ 0.0377\end{tabular} & \begin{tabular}[c]{@{}c@{}}0.36516 \\ $\pm$ 0.0048\end{tabular}                                                   & \begin{tabular}[c]{@{}c@{}}3.66753 \\ $\pm$ 0.0113\end{tabular}          & \begin{tabular}[c]{@{}c@{}}2.87278\\ $\pm$ 0.1182\end{tabular}                  & \begin{tabular}[c]{@{}c@{}}0.72683\\ $\pm$ 0.0282\end{tabular}                  \\ \hline
\begin{tabular}[c]{@{}c@{}}DLU\\ initialization\end{tabular}      & \begin{tabular}[c]{@{}c@{}}0.95001 \\ $\pm$ 0.0184\end{tabular} & \begin{tabular}[c]{@{}c@{}}0.32673 \\ $\pm$ 0.0052\end{tabular}                                                   & \begin{tabular}[c]{@{}c@{}}3.76655
 \\ $\pm$ 0.0052\end{tabular}           & \begin{tabular}[c]{@{}c@{}}02.82699\\ $\pm$ 0.0714\end{tabular}                  & \begin{tabular}[c]{@{}c@{}}0.69795\\ $\pm$ 0.0199\end{tabular}                  \\ \hline
\begin{tabular}[c]{@{}c@{}}Proposed\\ initialization\end{tabular} & \begin{tabular}[c]{@{}c@{}}0.899799 \\ $\pm$ 0.0210\end{tabular} & \begin{tabular}[c]{@{}c@{}}0.380161 \\ $\pm$ 0.0029\end{tabular}                                                   & \begin{tabular}[c]{@{}c@{}}\textbf{3.56722} \\ $\pm$ \textbf{0.0095}\end{tabular} & \begin{tabular}[c]{@{}c@{}}\textbf{3.10646}\\ $\pm$ \textbf{0.1017}\end{tabular}         & \begin{tabular}[c]{@{}c@{}}\textbf{0.74726}\\ $\pm$ \textbf{0.0226}\end{tabular}      
\end{tabular}
}
\end{table}

\subsubsection{Evaluation of different iterative algorithms}\label{ss:diff}
The proposed framework is designed to optimize the parameter set of any iterative reconstruction method. To demonstrate this, we applied the optimization process to the ASD-POCS and AwPCSD algorithms, which have 8 and 6 tunable parameters, respectively. These two algorithms were selected because they perform well with limited angle projection data, require more tunable parameters compared to others, and allow comparison with the brute-force method used in \cite{lohvithee2017parameter}. Additionally, to demonstrate the superiority of the proposed SSA-CSA over the original CSA, we conducted all experiments using the original CSA as well. 

\begin{figure}[]
	\centering
		\includegraphics[width=0.8\linewidth]{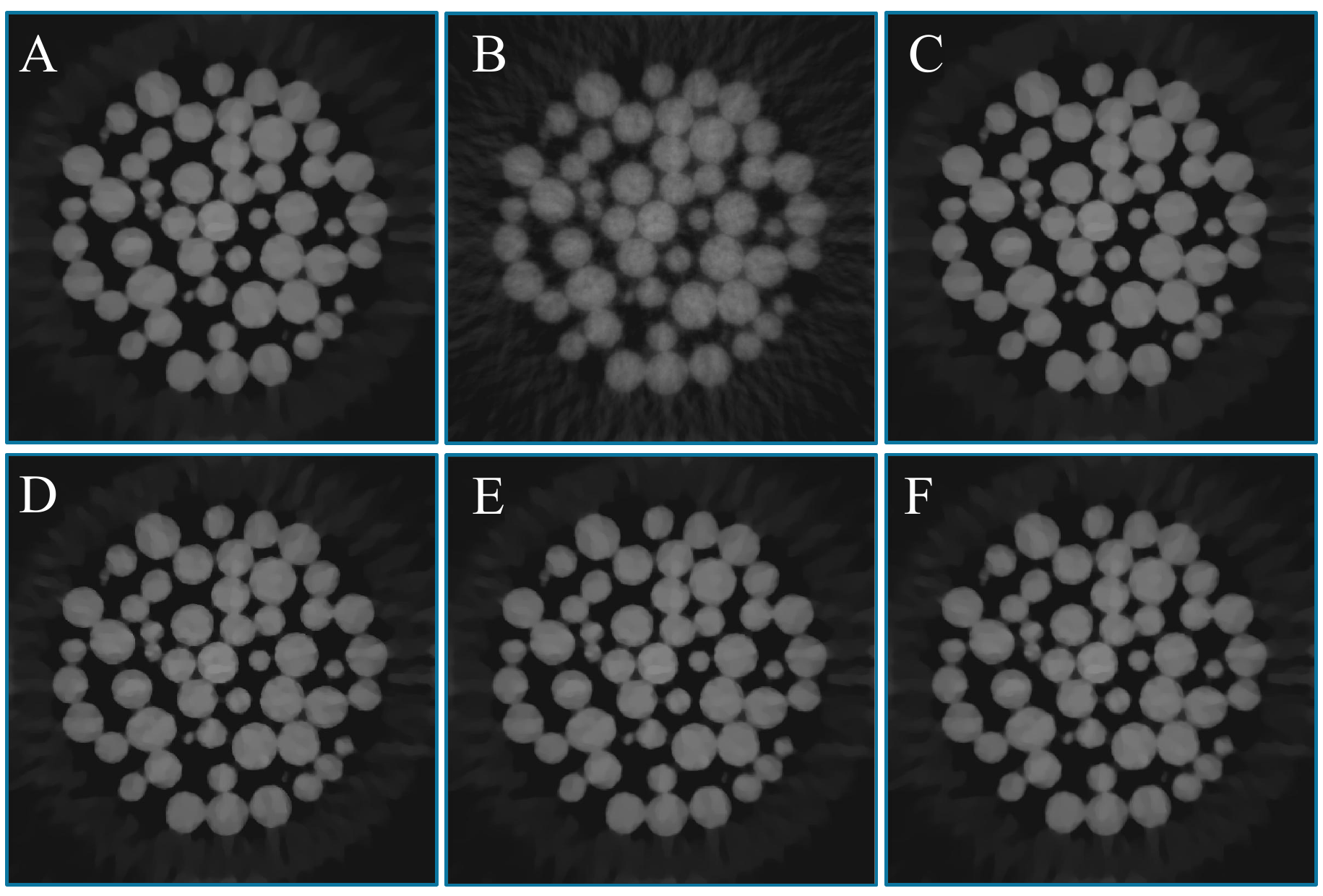}
	\caption{Qualitative results of Manual, CSA, and SSA-CSA on (A-C) ASD-POCS, and (E-F) AwPCSD reconstruction on SophiaBeads dataset}
	\label{FIG:6}
\end{figure}

Table \ref{Tab:4} presents the results of this experiment. Manual search performs similarly to the original CSA in both algorithms, while SSA-CSA has found a relatively better balance point and, consequently, a better fitness value. Additionally, the solution found by SSA-CSA achieves higher scores from learning-based solutions compared to the manual and CSA solutions. The corresponding qualitative results are shown in Figure \ref{FIG:6}.

\begin{table}[]
\centering
\caption{Performance of the different optimization methods with two different reconstruction algorithm on SophiaBead dataset.}
\label{Tab:4}
\resizebox{1\columnwidth}{!}{%
\begin{tabular}{c|c|ccc|cc}
\multicolumn{1}{l|}{}                                                                & \multicolumn{1}{l|}{}                            & \multicolumn{1}{l}{\textbf{}}   & \multicolumn{1}{l}{\textbf{Classic metrics}} & \multicolumn{1}{l|}{\textbf{}}           & \multicolumn{2}{l}{\textbf{Learning-Based Models}}                                  \\ \cline{3-7} 
\multicolumn{1}{l|}{\multirow{-2}{*}{}}                                              & \multicolumn{1}{l|}{\multirow{-2}{*}{Algorithm}} & SNR                             & HFER                             & Fitness                                  & CHILL@UK                                 & RPI\_AXIS                                \\ \hline
                                                                                     & ASD-POCS                                         & 0.73021                         & 0.35616                                      & 3.85591                                  & 3.00685                         & 0.70177                                  \\
\multirow{-2}{*}{\begin{tabular}[c]{@{}c@{}}Manual Setting\\ \cite{lohvithee2017parameter}\end{tabular}} & \cellcolor[HTML]{E8E8E8}AwPCSD                   & \cellcolor[HTML]{E8E8E8}0.73533 & \cellcolor[HTML]{E8E8E8}0.36053              & \cellcolor[HTML]{E8E8E8}3.82957          & \cellcolor[HTML]{E8E8E8}3.07818          & \cellcolor[HTML]{E8E8E8}0.69942          \\ \hline
                                                                                     & ASD-POCS                                         & 1.00030                         & 0.26501                                      & 4.00725                                  & 2.59235                                  & 0.65613                                  \\
\multirow{-2}{*}{CSA}                                                                & \cellcolor[HTML]{E8E8E8}AwPCSD                   & \cellcolor[HTML]{E8E8E8}0.75690 & \cellcolor[HTML]{E8E8E8}0.35570              & \cellcolor[HTML]{E8E8E8}3.82416          & \cellcolor[HTML]{E8E8E8}\textbf{3.25605}          & \cellcolor[HTML]{E8E8E8}0.70814          \\ \hline
                                                                                     & ASD-POCS                                         & 0.71330                         & 0.37350                                      & \textbf{3.80060}                         & \textbf{3.08410}                                  & \textbf{0.73592}                         \\
\multirow{-2}{*}{SSA-CSA}                                                            & \cellcolor[HTML]{E8E8E8}AwPCSD                   & \cellcolor[HTML]{E8E8E8}0.74074 & \cellcolor[HTML]{E8E8E8}0.36129              & \cellcolor[HTML]{E8E8E8}\textbf{3.81920} & \cellcolor[HTML]{E8E8E8}3.10410 & \cellcolor[HTML]{E8E8E8}\textbf{0.72592}
\end{tabular}
}
\end{table}

\subsubsection{Full-reference image quality assessment}
Although the main focus of this paper considers a real clinical setup, in which we lack a reference image to perform reference-based optimization, in this experiment, considering the fact that the SophiaBeads dataset has a gold standard, we assessed the quality of all ablation studies conducted so far using several full-reference image quality assessment (FR-IQA) methods to evaluate the performance of our proposed framework. FR-IQA methods can yield results that are easier to interpret when appropriate reference data is available \cite{zarb2010image, breger2024study}, and allow us further insights in our test setting here.

To this end, eight commonly used image quality metrics were computed, including Peak Signal-to-Noise Ratio (PSNR), Feature Similarity Index (FSIM) and its chromatic variant (FSIMc) \cite{zhang2011fsim}, the Haar wavelet–based perceptual similarity index with optimized settings for medical images ($HaarPSI_{MED}$) \cite{karner2025parameter}, the Visual Saliency–Induced Index (VSI) \cite{zhang2014vsi}, Visual Information Fidelity (VIF) \cite{sheikh2006image}, Deep Image Structure and Texture Similarity (DISTS) \cite{ding2020image}, and Learned Perceptual Image Patch Similarity (LPIPS) \cite{zhang2018unreasonable} adapted and tested for medical images \cite{breger2024study, breger2025study}.

Table \ref{Tab:8} presents a guide on how to interpret the results of each metric, along with all numerical evaluations from the ablation studies in the previous sections. To ensure a fair comparison across all results, only the parameter optimization on ASD-POCS is presented. The results indicate that the proposed framework not only performs well on no-reference metrics but also achieves strong performance on FR-IQA methods. Specifically, in the HaarPSI metric, which is mainly affected  by edges and structural distortions, the proposed method demonstrated clear superiority with a notable margin compared to the others.

\begin{table*}[]
\centering
\caption{Full-reference image quality assessment results on the SophiaBeads dataset using the ASD-POCS algorithm. In the value assessment row, $\Uparrow$ indicates higher is better, and $\Downarrow$ indicates lower is better.}
\label{Tab:8}
\resizebox{1\linewidth}{!}{%
\begin{tabular}{cc|cccccccc}
                                                         &                                                  & \multicolumn{8}{c}{\textbf{Reference-based metrics}}                                                                                                                                                                                                                           \\ \cline{3-10} 
                                                         &                                                  & PSNR                             & FSIM                            & FSIMc                           & $Haarpsi_{MED}$                         & VSI                             & VIF                             & DISTS                           & LPIPS                           \\ \hline
\multicolumn{2}{c|}{Range}                                                                                  & {[}0 - inf{]} dB                 & {[}0 - 1{]}                     & {[}0 - 1{]}                     & {[}0 - 1{]}                     & {[}0 - 1{]}                     & {[}0 - 1{]}                     & {[}0 - 1{]}                     & {[}0 - inf{]}                   \\
\multicolumn{2}{c|}{Value assessment}                                                                       & $\Uparrow$                             & $\Uparrow$                            & $\Uparrow$                            & $\Uparrow$                            & $\Uparrow$                            & $\Uparrow$                            & $\Downarrow$                           & $\Downarrow$                           \\ \hline
\multicolumn{1}{c|}{}                                    & SNR                                              & 28.06274                         & 0.82860                         & 0.81270                         & 0.22222                         & 0.98467                         & 0.11960                         & 0.37901                         & 0.39808                         \\
\multicolumn{1}{c|}{}                                    & \cellcolor[HTML]{E8E8E8}Laplacian                & \cellcolor[HTML]{E8E8E8}29.11234 & \cellcolor[HTML]{E8E8E8}0.83340 & \cellcolor[HTML]{E8E8E8}0.82150 & \cellcolor[HTML]{E8E8E8}0.22854 & \cellcolor[HTML]{E8E8E8}0.98900 & \cellcolor[HTML]{E8E8E8}0.23730 & \cellcolor[HTML]{E8E8E8}0.35384 & \cellcolor[HTML]{E8E8E8}0.35896 \\
\multicolumn{1}{c|}{}                                    & HFER                                        & 30.46449                         & 0.83930                         & 0.83140                         & 0.26012                         & 0.99179                         & 0.39130                         & 0.33722                         & 0.32110                         \\
\multicolumn{1}{c|}{\multirow{-4}{*}{Fitness}}           & \cellcolor[HTML]{E8E8E8}SNR + Laplacian          & \cellcolor[HTML]{E8E8E8}30.05906 & \cellcolor[HTML]{E8E8E8}0.83720 & \cellcolor[HTML]{E8E8E8}0.82790 & \cellcolor[HTML]{E8E8E8}0.25127 & \cellcolor[HTML]{E8E8E8}0.99103 & \cellcolor[HTML]{E8E8E8}0.34620 & \cellcolor[HTML]{E8E8E8}0.34531 & \cellcolor[HTML]{E8E8E8}0.30470 \\ \hline
\multicolumn{1}{c|}{}                                    & Random                                           & 33.12295                         & 0.87900                         & 0.85540                         & 0.55299                         & 0.99629                         & 0.54580                         & 0.31496                         & 0.14474                         \\
\multicolumn{1}{c|}{}                                    & \cellcolor[HTML]{E8E8E8}LHS                      & \cellcolor[HTML]{E8E8E8}29.82170 & \cellcolor[HTML]{E8E8E8}0.82820 & \cellcolor[HTML]{E8E8E8}0.78020 & \cellcolor[HTML]{E8E8E8}0.38777 & \cellcolor[HTML]{E8E8E8}0.99514 & \cellcolor[HTML]{E8E8E8}0.46990 & \cellcolor[HTML]{E8E8E8}0.32662 & \cellcolor[HTML]{E8E8E8}0.19108 \\
\multicolumn{1}{c|}{\multirow{-3}{*}{Initialization}}    & DLU                                              & 29.14888                         & 0.81510                         & 0.76340                         & 0.33898                         & 0.99453                         & 0.43390                         & 0.33203                         & 0.22001                         \\ \hline
\multicolumn{1}{c|}{}                                    & \cellcolor[HTML]{E8E8E8}Manual Setting \cite{lohvithee2017parameter} & \cellcolor[HTML]{E8E8E8}33.41629 & \cellcolor[HTML]{E8E8E8}0.87220 & \cellcolor[HTML]{E8E8E8}0.85520 & \cellcolor[HTML]{E8E8E8}0.55970 & \cellcolor[HTML]{E8E8E8}0.99707 & \cellcolor[HTML]{E8E8E8}0.67160 & \cellcolor[HTML]{E8E8E8}0.28401 & \cellcolor[HTML]{E8E8E8}0.09766 \\
\multicolumn{1}{c|}{\multirow{-2}{*}{Different methods}} & CSA                                              & \textbf{35.60611}                & 0.83510                         & 0.82440                         & 0.24031                         & 0.99028                         & 0.29530                         & 0.35005                         & 0.31889                         \\ \hline
\multicolumn{2}{c|}{Proposed method (SSA-CSA)}                                                              & 33.91926                         & \textbf{0.88370}                & \textbf{0.86580}                & \textbf{0.61606}                & \textbf{0.99724}                & \textbf{0.69910}                & \textbf{0.28084}                & \textbf{0.08530}               
\end{tabular}
}
\end{table*}

\subsection{Results on LinePairs CatPhan phantom}\label{ss:LinePaire}

\begin{figure}[]
	\centering
		\includegraphics[width=0.8\linewidth]{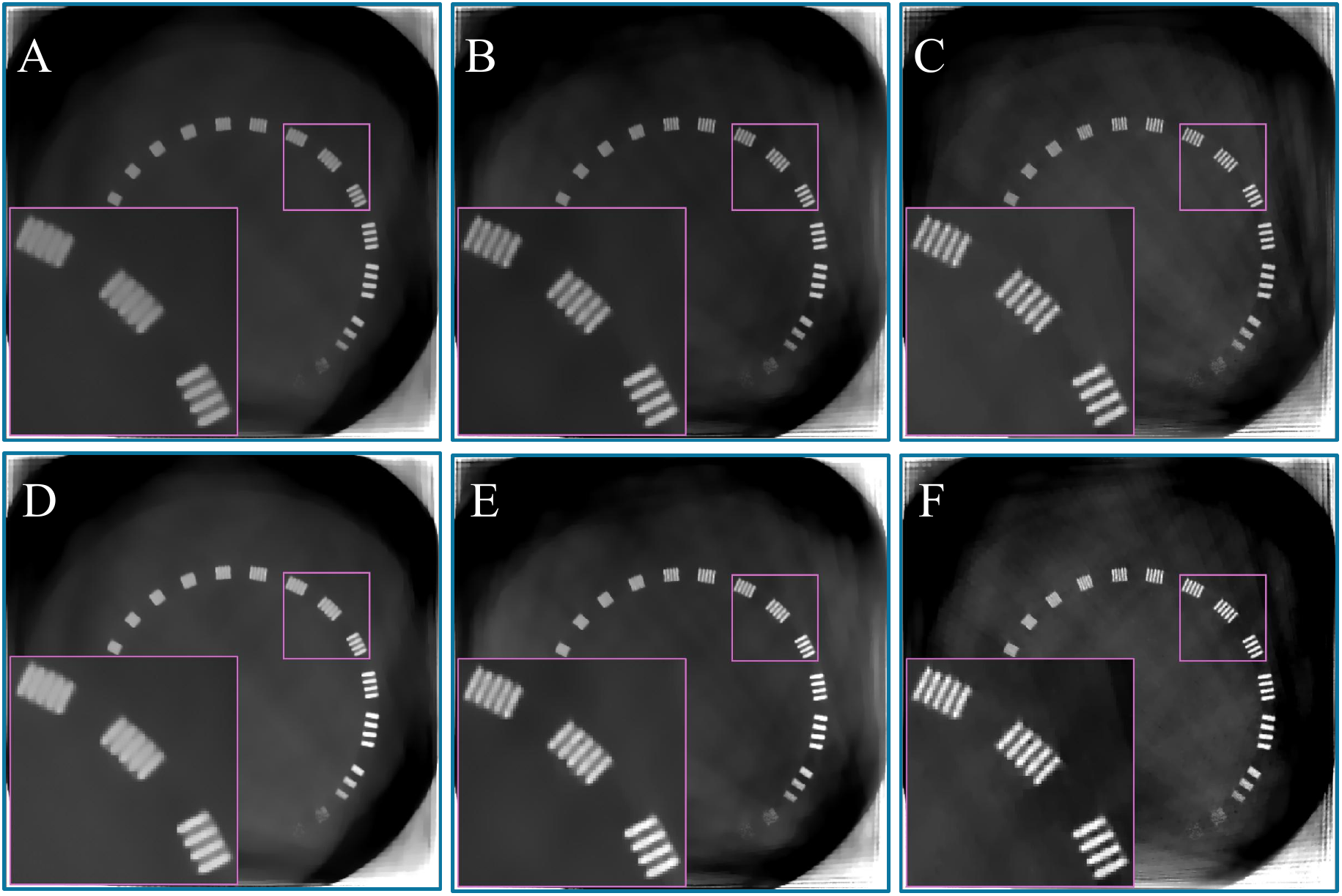}
	\caption{Qualitative results of Manual, CSA, and SSA-CSA on (A-C) ASD-POCS, and (E-F) AwPCSD reconstruction on LinePairs dataset}
	\label{FIG:7}
\end{figure}

The important aspect of reconstructing the line-pair phantom is the separability of the reconstructed lines. A very smooth image can distort the very thin boundaries, while a very sharp image may also reduce boundary visibility due to noise. We ran both CSA and SSA-CSA on the ASD-POCS and AwPCSD algorithms and compared the results with an manual parameter setting, selected through some practical trials to report the best outcome. Table \ref{Tab:5} and Figure \ref{FIG:7} present the quantitative and qualitative results, respectively.

\begin{table*}[!h]
\centering
\caption{Performance of the different optimization methods with two different reconstruction algorithm on LinePairs dataset.}
\label{Tab:5}
\resizebox{1\linewidth}{!}{%
\begin{tabular}{c|c|ccc|cc}
                                                                              &                                & \multicolumn{3}{c|}{Classic metrics}                                                                         & \multicolumn{2}{c}{Learning-Based Models}                         \\ \cline{3-7} 
\multirow{-2}{*}{}                                                            & \multirow{-2}{*}{Algorithm}    & SNR                             & HFER                & Fitness                                  & CHILL@UK                        & RPI\_AXIS                       \\ \hline
                                                                              & ASD-POCS                       & 2.19400                         & 0.42660                         & 2.89935                                  & 2.19328                         & 0.61135                         \\
\multirow{-2}{*}{\begin{tabular}[c]{@{}c@{}}Manual\\ setting\end{tabular}} & \cellcolor[HTML]{E8E8E8}AwPCSD & \cellcolor[HTML]{E8E8E8}1.67370 & \cellcolor[HTML]{E8E8E8}0.49310 & \cellcolor[HTML]{E8E8E8}2.69929          & \cellcolor[HTML]{E8E8E8}2.47366 & \cellcolor[HTML]{E8E8E8}0.62196 \\ \hline
                                                                              & ASD-POCS                       & 1.99470                         & 0.46820                         & 2.74403                                  & \textbf{2.53246}                         & 0.62230                         \\
\multirow{-2}{*}{CSA}                                                         & \cellcolor[HTML]{E8E8E8}AwPCSD & \cellcolor[HTML]{E8E8E8}1.64170 & \cellcolor[HTML]{E8E8E8}0.49730 & \cellcolor[HTML]{E8E8E8}2.68854          & \cellcolor[HTML]{E8E8E8}2.46662 & \cellcolor[HTML]{E8E8E8}0.61846 \\ \hline
                                                                              & ASD-POCS                       & 1.83100                         & 0.50060                         & \textbf{2.62960}                         & 2.48891                         & \textbf{0.63080}                         \\
\multirow{-2}{*}{SSA-CSA}                                                     & \cellcolor[HTML]{E8E8E8}AwPCSD & \cellcolor[HTML]{E8E8E8}1.61030 & \cellcolor[HTML]{E8E8E8}0.50820 & \cellcolor[HTML]{E8E8E8}\textbf{2.64780} & \cellcolor[HTML]{E8E8E8}\textbf{2.50507} & \cellcolor[HTML]{E8E8E8}\textbf{0.63758}
\end{tabular}
}
\end{table*}

From Figure \ref{FIG:7}, it is evident that the ASD-POCS algorithm shows substantial improvement when using the optimized parameter set obtained through SSA-CSA compared to other two methods. Moreover, the results for AwPCSD on this dataset show visually similar outcomes for CSA and SSA-CSA, which follows the same behavior as observed in table \ref{Tab:5}.

\begin{figure}[]
	\centering
		\includegraphics[width=0.8\linewidth]{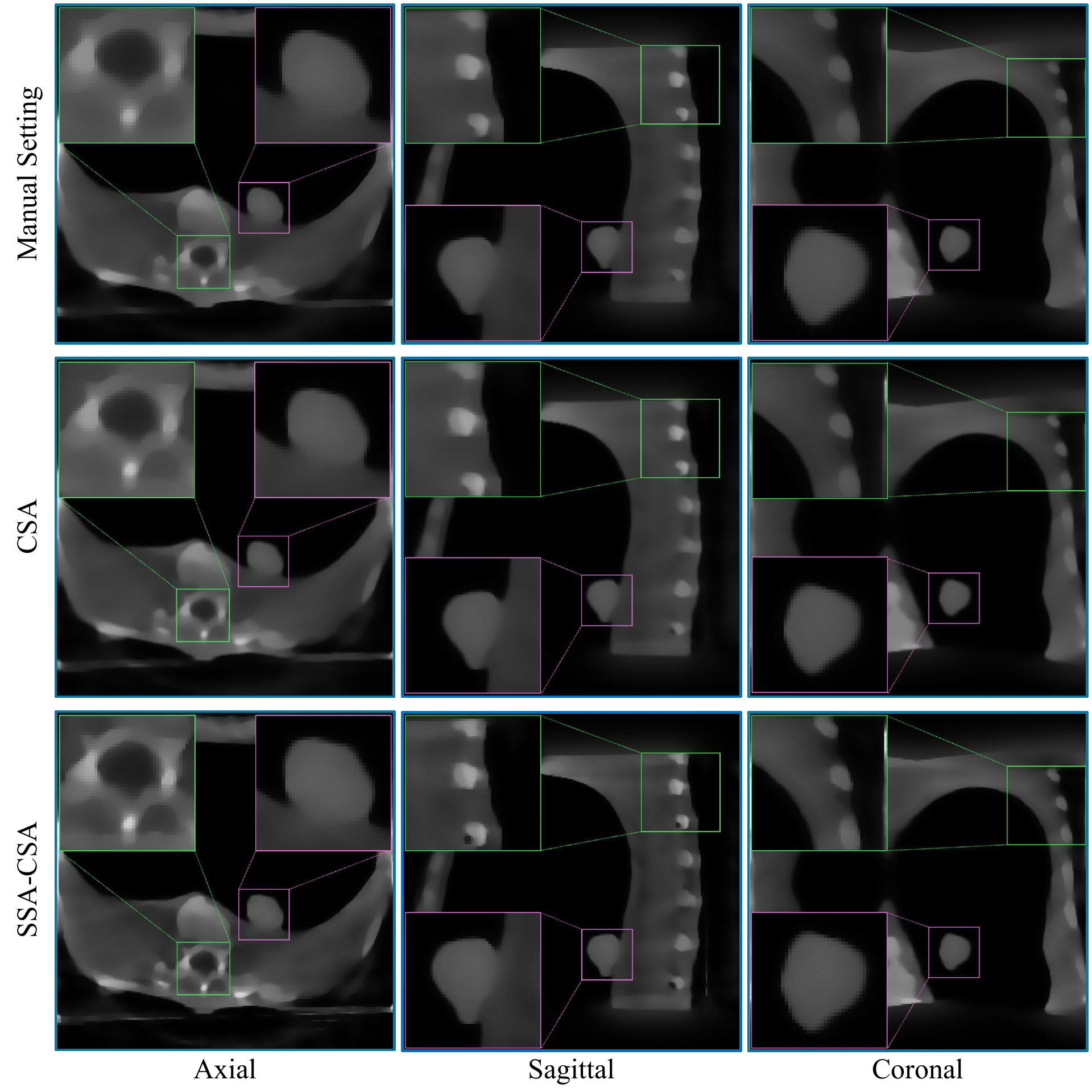}
	\caption{ASD-POCS reconstruction results on Thorax data using different optimization approaches}
	\label{FIG:8}
\end{figure}

\begin{figure}[]
	\centering
		\includegraphics[width=0.8\linewidth]{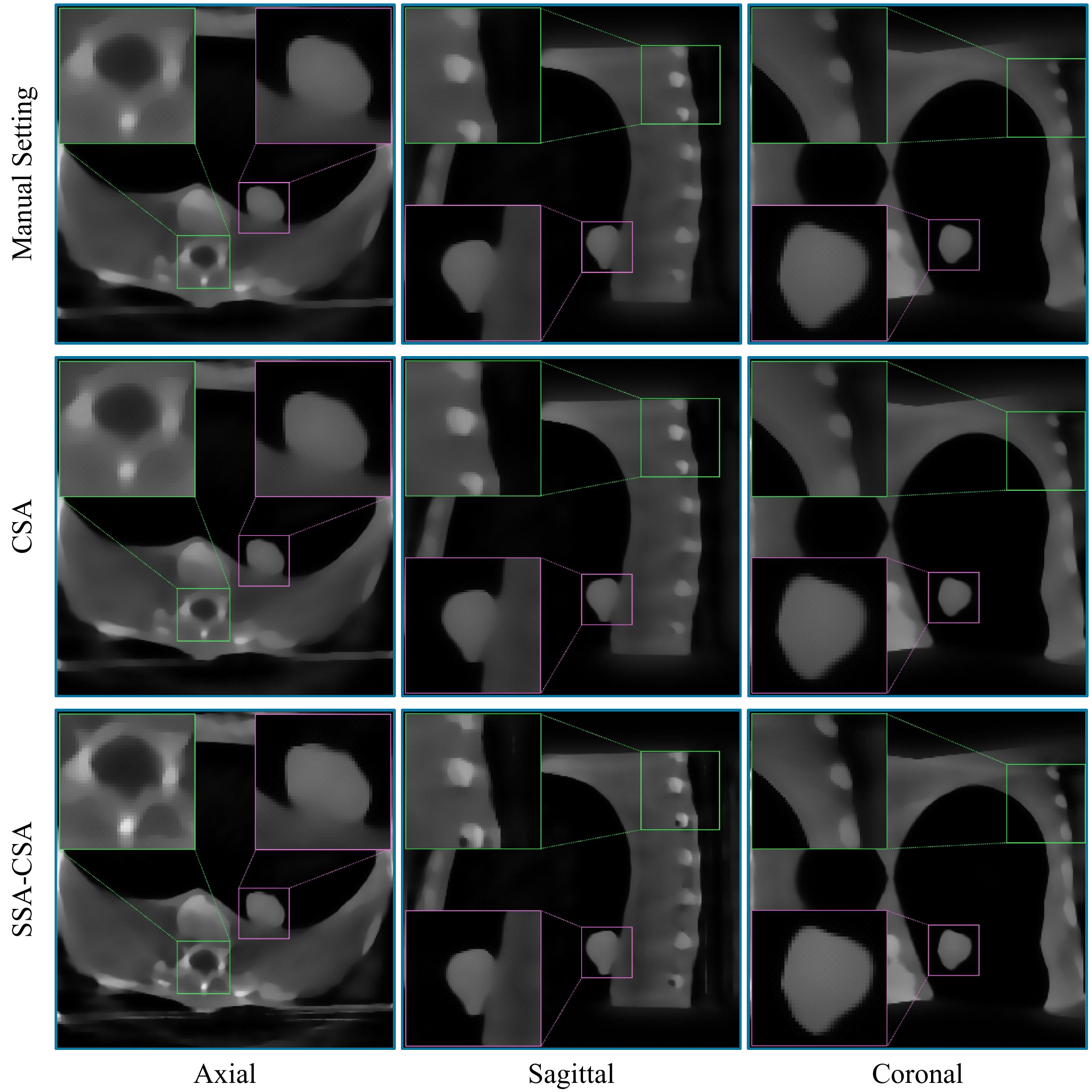}
	\caption{AwPCSD reconstruction results on Thorax data using different optimization approaches}
	\label{FIG:9}
\end{figure}

\subsection{Reconstruction results on Thorax phantom}\label{ss:thorax}
The Thorax data is an anatomical data of a printed phantom, which has a very high similarity to real patient skeletal structures. Therefore, the results on Thorax data reflect the real performance of the proposed framework in other clinical settings with limited projections. Similar to the experiments on the linepairs phantom, we conducted three optimization methods on two reconstruction algorithms to evaluate the results. Table \ref{Tab:6} presents classic and learning-based metrics for all experiments. The SSA-CSA successfully outperforms both manual and CSA optimization in the ASD-POCS and AwPCSD algorithms. For AwPCSD, the manual and CSA methods perform very similarly, but with the improvements made on CSA, the optimized parameter set found by SSA-CSA achieves better results.
\begin{table*}[]
\centering
\caption{Performance of the different optimization methods with two different reconstruction algorithm on Thorax dataset.}
\label{Tab:6}
\resizebox{1\linewidth}{!}{%
\begin{tabular}{c|c|ccc|cc}
\multicolumn{1}{l|}{}                                                                & \multicolumn{1}{l|}{}                            & \multicolumn{1}{l}{\textbf{}}   & \multicolumn{1}{l}{\textbf{Classic metrics}}              & \multicolumn{1}{l|}{\textbf{}}           & \multicolumn{2}{l}{\textbf{Learning-Based Models}}                                  \\ \cline{3-7} 
\multicolumn{1}{l|}{\multirow{-2}{*}{}}                                              & \multicolumn{1}{l|}{\multirow{-2}{*}{Algorithm}} & SNR                             & HFER & Fitness                                  & CHILL@UK                                 & RPI\_AXIS                                \\ \hline
                                                                                     & ASD-POCS                                         & 1.26260                         & 0.31520                                                   & 3.63601                                  & 2.61885                                  & 0.60266                                  \\
\multirow{-2}{*}{\begin{tabular}[c]{@{}c@{}}Manual\\ Setting\end{tabular}} & \cellcolor[HTML]{E8E8E8}AwPCSD                   & \cellcolor[HTML]{E8E8E8}1.28770 & \cellcolor[HTML]{E8E8E8}0.29850                           & \cellcolor[HTML]{E8E8E8}3.70035          & \cellcolor[HTML]{E8E8E8}2.42057          & \cellcolor[HTML]{E8E8E8}0.58130          \\ \hline
                                                                                     & ASD-POCS                                         & 1.22730                         & 0.32620                                                   & 3.60246                                  & 2.57055                                  & 0.58787                                  \\
\multirow{-2}{*}{CSA}                                                                & \cellcolor[HTML]{E8E8E8}AwPCSD                   & \cellcolor[HTML]{E8E8E8}1.24460 & \cellcolor[HTML]{E8E8E8}0.30340                           & \cellcolor[HTML]{E8E8E8}3.69713          & \cellcolor[HTML]{E8E8E8}2.62096          & \cellcolor[HTML]{E8E8E8}0.57254          \\ \hline
                                                                                     & ASD-POCS                                         & 1.10710                         & 0.37380                                                   & \textbf{3.45018}                         & \textbf{2.74691}                         & \textbf{0.63154}                         \\
\multirow{-2}{*}{SSA-CSA}                                                            & \cellcolor[HTML]{E8E8E8}AwPCSD                   & \cellcolor[HTML]{E8E8E8}1.13640 & \cellcolor[HTML]{E8E8E8}0.34290                           & \cellcolor[HTML]{E8E8E8}\textbf{3.57293} & \cellcolor[HTML]{E8E8E8}\textbf{2.64249} & \cellcolor[HTML]{E8E8E8}\textbf{0.59609}
\end{tabular}
}
\end{table*}

As Thorax data is anatomical, it is important to investigate the qualitative results across all views. Figures \ref{FIG:8} and \ref{FIG:9} present the qualitative results of the three applied optimization methods on the ASD-POCS and AwPCSD algorithms, respectively. The qualitative results show that SSA-CSA found the hyperparameters in a way that preserves sharp edges while maintaining a uniform texture in the inner structures of the organs.



\subsection{Results on brain phantom dataset}\label{ss:brain}

In this section, the results of applying the PICCS algorithm to brain phantom data are presented. In this experiment, to ensure a fair comparison between manual setting, CSA, and the proposed SSA-CSA, we used the same prior knowledge to run PICCS during the algorithm run  and for the final solution reconstruction. Additionally, as mentioned in the dataset description in Section \ref{Brain_data_Sec}, to simulate a clinical scenario, we used only 20\% of the total 473 projections for the post-operative brain data after screw insertion. A major challenge with this data is the very small details in the texture of the phantom, such as small holes or structures mimicking the veins in the brain, which means that even minor errors in the parameter settings of PICCS can result in a reconstruction that loses these fine details. 

\begin{figure}[]
	\centering
		\includegraphics[width=1\linewidth]{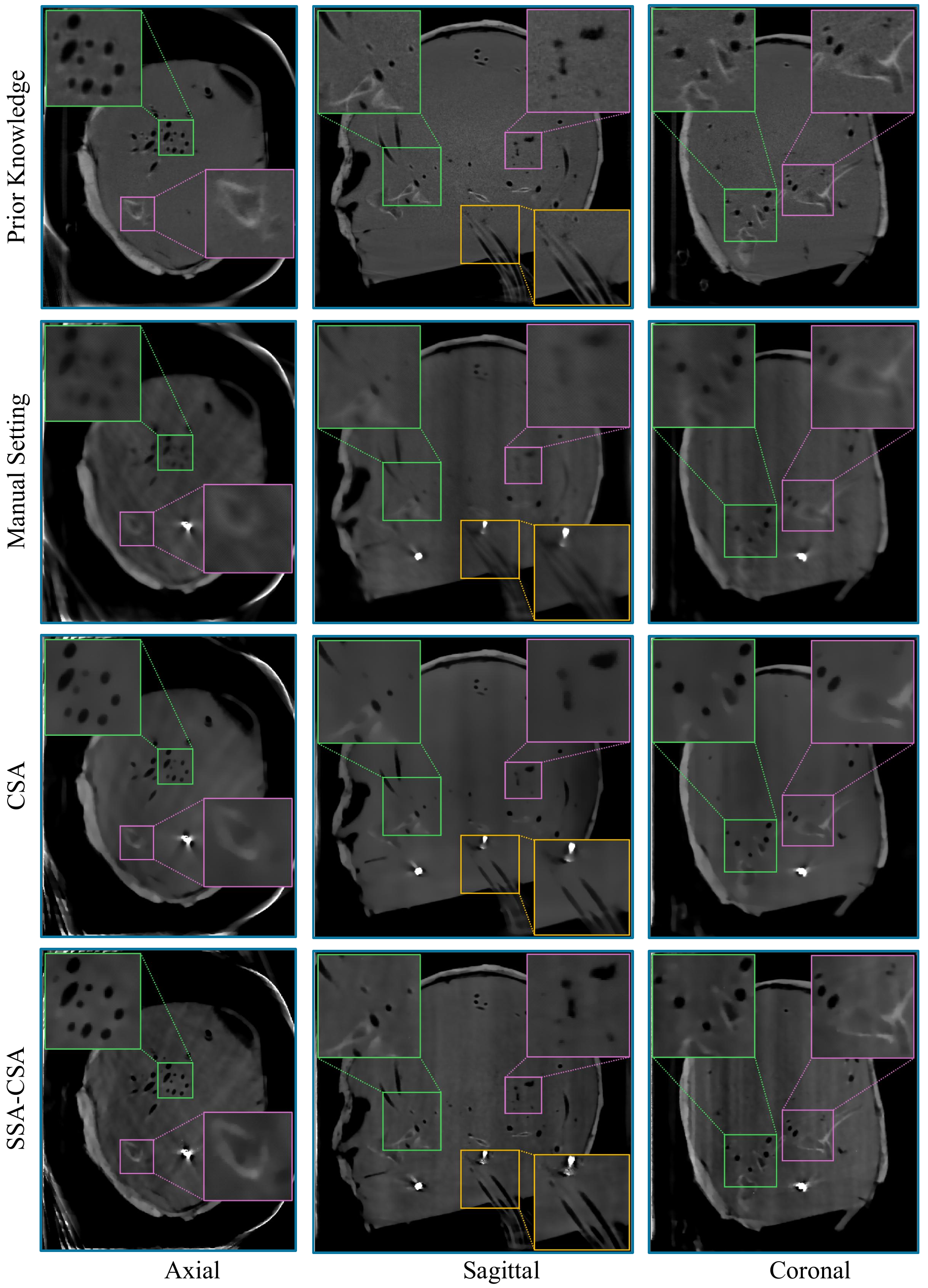}
	\caption{PICCS reconstruction results on Brain data using different optimization approaches. First row indicates the prior knowledge for all algorithms.}
	\label{FIG:12}
\end{figure}

The results in Table \ref{Tab:9} clearly show that SSA-CSA could find the best balance between SNR and HFER compared to the other methods and is superior across all metrics. To gain a better understanding of the presented quantitative results, as illustrated in Figure \ref{FIG:12}, it is evident that SSA-CSA could successfully find a parameter set that maintains fine details sharply while also providing a clear presentation of the inserted metal screw in the brain. On the other hand, CSA could not find a solution that preserves these details as effectively as SSA-CSA, and finally, the manual setting results show that in some regions the details are completely merged with the surrounding area due to over-smoothing.

\begin{table}[]
\centering
\caption{Performance of different optimization methods with PICCS algorithm on brain phantom}
\label{Tab:9}
\resizebox{0.8\linewidth}{!}{%
\begin{tabular}{c|ccc|cc}
\multicolumn{1}{l|}{\multirow{2}{*}{}} & \multicolumn{3}{c|}{\textbf{Classic metrics}}          & \multicolumn{2}{c}{\textbf{Learning-Based Models}} \\ \cline{2-6} 
\multicolumn{1}{l|}{}                  & SNR     & HFER & Fitness          & CHILL@UK            & RPI\_AXIS           \\ \hline
\begin{tabular}[c]{@{}c@{}}Manual\\ setting\end{tabular}                   & 1.69057 & 0.33843          & 3.39111          & 2.42896             & 0.66923             \\ \hline
CSA                                    & 1.53696 & 0.40155          & 3.14846          & 2.62561             & 0.69504             \\ \hline
SSA-CSA                                & 1.28064 & 0.44485          & \textbf{3.04477} & \textbf{2.67673}    & \textbf{0.70178}   
\end{tabular}
}
\end{table}

\section{Discussion} \label{S:diss}

This research introduced a general and scalable framework that can be used as a tool to better leverage iterative image reconstruction methods in CBCT by identifying an optimal parameter set specific to the algorithm and the dataset. An in-depth evaluation was performed to assess all different aspects of the proposed SSA-CSA. Based on extensive ablation studies and experiments on four different real-world datasets, the method is able to identify a high-quality parameter set within a very large search space that is not feasible using brute-force strategies or manual trial-and-error approaches.

As we can see in all results tables, the metrics are very close to each other, especially the classical metrics. This is because the ranges of these metrics are bounded. Even for SNR, which theoretically has no upper bound, in reconstructed medical images of this study it is practically limited to a range of 0.7 to 2.5. This makes the numerical results appear very close. However, even small differences in these metrics indicate some minor structural variations in the reconstructed image, which can be clinically significant and based on the results presented in Table \ref{Tab:8}, it can be observed that the proposed method also achieves better results in FR-IQA methods. 


As discussed in section \ref{S:relatedworks}, the method presented in \cite{lohvithee2021ant} is the closest state-of-the-art approach to the present research. However, due to differences in the experimental setup such as the use of different dataset and the unavailability of the original code a fair and comprehensive comparison with that work was not feasible. Nevertheless, based on the challenges outlined in the text, which SSA-CSA successfully addresses, no corresponding efforts were made in the related work to tackle these issues. In particular, the utilized metaheuristic, (ACO) \cite{dorigo2007ant}, was not further enhanced, progressive narrowing of the search space was applied through parameter freezing or limiting the range of tunable parameters, and most importantly, their approach relied on a reference-based fitness function. These limitations further highlight the methodological superiority of SSA-CSA in the context of real-world clinical applications.

Although the proposed method can efficiently identify a good hyperparameter set for the reconstruction methods, it also has some drawbacks. The first point is its metaheuristic nature; While it can explore a large search space efficiently, like all metaheuristics, it cannot guarantee finding the best global optimum and may instead converge to a local optimum. Another limitation arises from the combined fitness formulation, which requires tuning two weights. On the one hand, these weights provide users with greater flexibility to adjust the focus of the algorithm; on the other hand, in some cases, setting these weights may be challenging.

\section{Conclusion} \label{S:conc}
In this work, we tried to leverage the benefits of iterative reconstruction algorithms by optimizing the hyperparameter set of these algorithms. Since there are many possible values for each parameter, finding the optimal set through a brute-force search is impractical. Hence, by utilizing a modified version of the bio-inspired CSA, we tried to address this optimization problem. In the proposed SSA-CSA, we incorporated a more efficient initialization approach, a more goal-oriented local search, and a search-space-aware global search mechanism. In addition, to maintain the balance between exploitation and exploration, we applied a balancing technique based on the previous performance of the crows. To set a goal for the algorithm and evaluate the performance of individuals during the process, a multi-objective no-reference fitness function was defined, considering both noise reduction and edge preservation.

The results demonstrated superiority over manual parameter setting and the original CSA across all analyses, outperforming them by improving the average fitness by 4.19\%, and achieving improvements of 4.89\% and 3.82\% on the CHILL@UK and RPI\_AXIS metrics, respectively. In addition, a high correlation between the proposed fitness function and other learning-based image quality assessment models was observed, indicating that despite its computational simplicity, it is representative of image quality. A future direction of this work would be incorporating a deep learning model as the fitness function in a way that it can be trained and predicts the corresponding fitness score. Moreover, as deep reinforcement learning techniques have shown promising results in similar applications, employing such methods instead of the metaheuristic component represents another potential extension of this work.

\section*{Acknowledgement}
We acknowledge the use of the AI-based language model ChatGPT (GPT-4.1, OpenAI) for grammar checking during the final stage of manuscript preparation. The authors take full responsibility for the content of the published article.

This study was funded by Austrian Research Promotion Agency (FFG) (FFG number: F0999914792).

\bibliographystyle{model1-num-names}

\bibliography{cas-refs}

\newpage
\section*{\textcolor{blue}{Appendix}}

\subsection*{\textcolor{blue}{A.} Results correlation analysis}\label{App_A}
According to the results of tables \ref{Tab:4}, \ref{Tab:5}, and \ref{Tab:6}, one notable finding is the relation between the proposed fitness function and the deep learning models used to assess the quality of reconstructed images. To better understand this relation, the Pearson correlation is calculated between the fitness function and each of the deep learning model predictions using Eq. \ref{Eq:10}.

\begin{equation}
\centering 
\label{Eq:10}
r(X, Y) = \frac{\sum_{i=1}^{n}(X_i - \bar{X})(Y_i - \bar{Y})}{\sqrt{\sum_{i=1}^{n}(X_i - \bar{X})^2}.\sqrt{\sum_{i=1}^{n}(Y_i - \bar{Y})^2}}
\end{equation}

Where $X_i$ and $Y_i$ are fitness and deep learning model scores, and $\bar{X}$ and $\bar{Y}$ are the means of the two sets. The Pearson correlation ranges from $-$1 to +1, indicating perfect inverse and direct correlation, respectively, while a value of 0 indicates no correlation.

Table \ref{Tab:7} presents the correlation of the fitness function with three learning-based models, calculated separately for all experiments. It can be seen that the proposed fitness has a strong inverse correlation with both CHILL@UK and RPI\_AXIS, which is expected to be inverse since the fitness is defined in minimization form.

\begin{table}[!h]
\centering
\caption{Pearson correlation of proposed fitness and different learning-based quality assessments for different datasets.}
\label{Tab:7}
\resizebox{0.7\columnwidth}{!}{%
\begin{tabular}{c|cc}
\multirow{2}{*}{}                                         & \multicolumn{2}{c}{Fitness Correlation} \\ \cline{2-3} 
                                                          & CHILL@UK           & RPI\_AXIS          \\ \hline
SophiaBeads                                               & -0.93758           & -0.94000           \\ \hline
\begin{tabular}[c]{@{}c@{}}LinePairs\end{tabular} & -0.86564           & -0.81844           \\ \hline
Thorax data                                               & -0.77732           & -0.91589          \\ \hline
Brain data                                               & -0.99507
           & -0.99516
\end{tabular}
}
\end{table}

\subsection*{\textcolor{blue}{B.} Hyperparameter values of experiments}\label{App_B}

In this section, the final solutions of all experiments performed in this paper are presented in table \ref{Tab:10}. This table facilitates the analysis of exact parameter values and highlights the importance of tuning, as in some cases the solution values are similar, but minor changes can lead to notable differences in the reconstructed image. Additionally, this table supports the reproducibility of the results, especially for the SophiaBeads dataset, which is publicly available in the literature. Providing the complete parameter set enables fair comparisons for future work on similar parameter optimization problems for iterative reconstruction alogorithms.

\begin{landscape}
\begin{table}[]
\centering
\caption{Final solution of all performed experiments with different scenarios and setups}
\label{Tab:10}
\resizebox{1.1\linewidth}{!}{%
\begin{tabular}{c|c|c|ccccccccc}
\textbf{Section} &
  \textbf{Algorithm} &
  \textbf{Experiment} &
  \textbf{$Max\_iter$} &
  \textbf{$TV\_iter$} &
  \textbf{$\epsilon$} &
  \textbf{$\alpha$} &
  \textbf{$\alpha_{red}$} &
  \textbf{$\lambda$} &
  \textbf{$\lambda_{red}$} &
  \textbf{$r\_max$} &
  \textbf{$\delta$} \\ \hline
 &
   &
  SNR &
  5 &
  6 &
  1072 &
  0.1000 &
  0.950 &
  0.924 &
  0.910 &
  0.900 &
  - \\
 &
   &
  \cellcolor[HTML]{E8E8E8}Laplacian &
  \cellcolor[HTML]{E8E8E8}8 &
  \cellcolor[HTML]{E8E8E8}12 &
  \cellcolor[HTML]{E8E8E8}83 &
  \cellcolor[HTML]{E8E8E8}0.0002 &
  \cellcolor[HTML]{E8E8E8}0.940 &
  \cellcolor[HTML]{E8E8E8}0.997 &
  \cellcolor[HTML]{E8E8E8}0.908 &
  \cellcolor[HTML]{E8E8E8}0.980 &
  \cellcolor[HTML]{E8E8E8}- \\
 &
   &
  HFER &
  15 &
  10 &
  56 &
  0.0001 &
  0.997 &
  0.971 &
  0.938 &
  0.900 &
  - \\
\multirow{-4}{*}{\textbf{\begin{tabular}[c]{@{}c@{}}\ref{ss:fitness}\\ Fitness\\ functions\end{tabular}}} &
  \multirow{-4}{*}{\textbf{ASD-POCS}} &
  \cellcolor[HTML]{E8E8E8}SNR + Laplacian &
  \cellcolor[HTML]{E8E8E8}48 &
  \cellcolor[HTML]{E8E8E8}9 &
  \cellcolor[HTML]{E8E8E8}309 &
  \cellcolor[HTML]{E8E8E8}0.0001 &
  \cellcolor[HTML]{E8E8E8}0.975 &
  \cellcolor[HTML]{E8E8E8}0.916 &
  \cellcolor[HTML]{E8E8E8}0.929 &
  \cellcolor[HTML]{E8E8E8}0.900 &
  \cellcolor[HTML]{E8E8E8}- \\ \hline
 &
   &
  Random &
  50 &
  25 &
  347 &
  0.0217 &
  0.934 &
  0.999 &
  0.944 &
  0.904 &
  - \\
 &
   &
  \cellcolor[HTML]{E8E8E8}LHS &
  \cellcolor[HTML]{E8E8E8}42 &
  \cellcolor[HTML]{E8E8E8}20 &
  \cellcolor[HTML]{E8E8E8}983 &
  \cellcolor[HTML]{E8E8E8}0.0135 &
  \cellcolor[HTML]{E8E8E8}0.956 &
  \cellcolor[HTML]{E8E8E8}0.983 &
  \cellcolor[HTML]{E8E8E8}0.967 &
  \cellcolor[HTML]{E8E8E8}0.950 &
  \cellcolor[HTML]{E8E8E8}- \\
\multirow{-3}{*}{\textbf{\begin{tabular}[c]{@{}c@{}}\ref{ss:init}\\ Initialization\\ methods\end{tabular}}} &
  \multirow{-3}{*}{\textbf{ASD-POCS}} &
  DLU &
  41 &
  19 &
  660 &
  0.0800 &
  0.990 &
  0.993 &
  0.930 &
  0.910 &
  - \\ \hline
 &
  \textbf{ASD-POCS} &
  \cellcolor[HTML]{E8E8E8}Manual setting \cite{lohvithee2017parameter} &
  \cellcolor[HTML]{E8E8E8}50 &
  \cellcolor[HTML]{E8E8E8}25 &
  \cellcolor[HTML]{E8E8E8}1500 &
  \cellcolor[HTML]{E8E8E8}0.0020 &
  \cellcolor[HTML]{E8E8E8}0.900 &
  \cellcolor[HTML]{E8E8E8}0.999 &
  \cellcolor[HTML]{E8E8E8}0.990 &
  \cellcolor[HTML]{E8E8E8}0.990 &
  \cellcolor[HTML]{E8E8E8}- \\
 &
  \textbf{AwPCSD} &
  Manual setting \cite{lohvithee2017parameter} &
  50 &
  6 &
  1500 &
  - &
  - &
  0.999 &
  0.990 &
  - &
  0.922 \\ \cline{2-12} 
 &
  \textbf{ASD-POCS} &
  \cellcolor[HTML]{E8E8E8}CSA &
  \cellcolor[HTML]{E8E8E8}16 &
  \cellcolor[HTML]{E8E8E8}22 &
  \cellcolor[HTML]{E8E8E8}321 &
  \cellcolor[HTML]{E8E8E8}0.0001 &
  \cellcolor[HTML]{E8E8E8}0.980 &
  \cellcolor[HTML]{E8E8E8}0.920 &
  \cellcolor[HTML]{E8E8E8}0.971 &
  \cellcolor[HTML]{E8E8E8}0.950 &
  \cellcolor[HTML]{E8E8E8}- \\
\multirow{-4}{*}{\textbf{\begin{tabular}[c]{@{}c@{}}\ref{ss:diff}\\ Different\\ algorithms\end{tabular}}} &
  \textbf{AwPCSD} &
  CSA &
  44 &
  25 &
  314 &
  - &
  - &
  0.960 &
  0.991 &
  - &
  0.360 \\ \hline
 &
  \textbf{ASD-POCS} &
  \cellcolor[HTML]{DDFFDD}Proposed method &
  \cellcolor[HTML]{DDFFDD}\textbf{49} &
  \cellcolor[HTML]{DDFFDD}\textbf{24} &
  \cellcolor[HTML]{DDFFDD}\textbf{698} &
  \cellcolor[HTML]{DDFFDD}\textbf{0.0032} &
  \cellcolor[HTML]{DDFFDD}\textbf{0.969} &
  \cellcolor[HTML]{DDFFDD}\textbf{0.998} &
  \cellcolor[HTML]{DDFFDD}\textbf{0.927} &
  \cellcolor[HTML]{DDFFDD}\textbf{0.974} &
  \cellcolor[HTML]{DDFFDD}\textbf{-} \\
\multirow{-2}{*}{\textbf{\begin{tabular}[c]{@{}c@{}}SSA-CSA solution\\for SophiaBeads\end{tabular}}} &
  \textbf{AwPCSD} &
  \cellcolor[HTML]{DDFFDD}Proposed method &
  \cellcolor[HTML]{DDFFDD}\textbf{45} &
  \cellcolor[HTML]{DDFFDD}\textbf{20} &
  \cellcolor[HTML]{DDFFDD}\textbf{810} &
  \cellcolor[HTML]{DDFFDD}\textbf{-} &
  \cellcolor[HTML]{DDFFDD}\textbf{-} &
  \cellcolor[HTML]{DDFFDD}\textbf{0.978} &
  \cellcolor[HTML]{DDFFDD}\textbf{0.982} &
  \cellcolor[HTML]{DDFFDD}\textbf{-} &
  \cellcolor[HTML]{DDFFDD}\textbf{0.244} \\ \hline
 &
   &
  Manual setting &
  20 &
  30 &
  1082 &
  0.0084 &
  0.957 &
  0.960 &
  0.922 &
  0.918 &
  - \\
 &
   &
  \cellcolor[HTML]{E8E8E8}CSA &
  \cellcolor[HTML]{E8E8E8}25 &
  \cellcolor[HTML]{E8E8E8}36 &
  \cellcolor[HTML]{E8E8E8}990 &
  \cellcolor[HTML]{E8E8E8}0.0076 &
  \cellcolor[HTML]{E8E8E8}0.981 &
  \cellcolor[HTML]{E8E8E8}0.983 &
  \cellcolor[HTML]{E8E8E8}0.930 &
  \cellcolor[HTML]{E8E8E8}0.923 &
  \cellcolor[HTML]{E8E8E8}- \\
 &
  \multirow{-3}{*}{\textbf{ASD-POCS}} &
   \cellcolor[HTML]{DDFFDD}Proposed method &
   \cellcolor[HTML]{DDFFDD}\textbf{34} &
   \cellcolor[HTML]{DDFFDD}\textbf{46} &
   \cellcolor[HTML]{DDFFDD}\textbf{650} &
   \cellcolor[HTML]{DDFFDD}\textbf{0.0015} &
   \cellcolor[HTML]{DDFFDD}\textbf{0.978} &
   \cellcolor[HTML]{DDFFDD}\textbf{0.999} &
   \cellcolor[HTML]{DDFFDD}\textbf{0.939} &
   \cellcolor[HTML]{DDFFDD}\textbf{0.980} &
   \cellcolor[HTML]{DDFFDD}\textbf{-} \\ \cline{2-12} 
 &
   &
  \cellcolor[HTML]{E8E8E8}Manual setting &
  \cellcolor[HTML]{E8E8E8}33 &
  \cellcolor[HTML]{E8E8E8}5 &
  \cellcolor[HTML]{E8E8E8}741 &
  \cellcolor[HTML]{E8E8E8}- &
  \cellcolor[HTML]{E8E8E8}- &
  \cellcolor[HTML]{E8E8E8}0.946 &
  \cellcolor[HTML]{E8E8E8}0.923 &
  \cellcolor[HTML]{E8E8E8}- &
  \cellcolor[HTML]{E8E8E8}0.051 \\
 &
   &
  CSA &
  13 &
  14 &
  1205 &
  - &
  - &
  0.952 &
  0.952 &
  - &
  0.063 \\
\multirow{-6}{*}{\textbf{\begin{tabular}[c]{@{}c@{}}\ref{ss:LinePaire}\\ Results on\\ LinePairs\\ phantom\end{tabular}}} &
  \multirow{-3}{*}{\textbf{AwPCSD}} &
  \cellcolor[HTML]{DDFFDD}Proposed method &
  \cellcolor[HTML]{DDFFDD}\textbf{23} &
  \cellcolor[HTML]{DDFFDD}\textbf{10} &
  \cellcolor[HTML]{DDFFDD}\textbf{430} &
  \cellcolor[HTML]{DDFFDD}\textbf{-} &
  \cellcolor[HTML]{DDFFDD}\textbf{-} &
  \cellcolor[HTML]{DDFFDD}\textbf{0.960} &
  \cellcolor[HTML]{DDFFDD}\textbf{0.980} &
  \cellcolor[HTML]{DDFFDD}\textbf{-} &
  \cellcolor[HTML]{DDFFDD}\textbf{0.065} \\ \hline
 &
   &
  Manual setting &
  43 &
  29 &
  692 &
  0.0999 &
  0.900 &
  0.980 &
  0.902 &
  0.960 &
  - \\
 &
   &
  \cellcolor[HTML]{E8E8E8}CSA &
  \cellcolor[HTML]{E8E8E8}34 &
  \cellcolor[HTML]{E8E8E8}43 &
  \cellcolor[HTML]{E8E8E8}1490 &
  \cellcolor[HTML]{E8E8E8}0.0291 &
  \cellcolor[HTML]{E8E8E8}0.948 &
  \cellcolor[HTML]{E8E8E8}0.967 &
  \cellcolor[HTML]{E8E8E8}0.931 &
  \cellcolor[HTML]{E8E8E8}0.934 &
  \cellcolor[HTML]{E8E8E8}- \\
 &
  \multirow{-3}{*}{\textbf{ASD-POCS}} &
  \cellcolor[HTML]{DDFFDD}Proposed method &
  \cellcolor[HTML]{DDFFDD}\textbf{37} &
  \cellcolor[HTML]{DDFFDD}\textbf{21} &
  \cellcolor[HTML]{DDFFDD}\textbf{1500} &
  \cellcolor[HTML]{DDFFDD}\textbf{0.0384} &
  \cellcolor[HTML]{DDFFDD}\textbf{0.937} &
  \cellcolor[HTML]{DDFFDD}\textbf{0.985} &
  \cellcolor[HTML]{DDFFDD}\textbf{0.962} &
  \cellcolor[HTML]{DDFFDD}\textbf{0.968} &
  \cellcolor[HTML]{DDFFDD}\textbf{-} \\ \cline{2-12} 
 &
   &
  Manual setting &
  16 &
  45 &
  200 &
  - &
  - &
  0.982 &
  0.941 &
  - &
  1.630 \\
 &
   &
  \cellcolor[HTML]{E8E8E8}CSA &
  \cellcolor[HTML]{E8E8E8}22 &
  \cellcolor[HTML]{E8E8E8}34 &
  \cellcolor[HTML]{E8E8E8}370 &
  \cellcolor[HTML]{E8E8E8}- &
  \cellcolor[HTML]{E8E8E8}- &
  \cellcolor[HTML]{E8E8E8}0.942 &
  \cellcolor[HTML]{E8E8E8}0.940 &
  \cellcolor[HTML]{E8E8E8}- &
  \cellcolor[HTML]{E8E8E8}1.540 \\
\multirow{-6}{*}{\textbf{\begin{tabular}[c]{@{}c@{}}\ref{ss:thorax}\\ Results on\\ Thorax\\ phantom\end{tabular}}} &
  \multirow{-3}{*}{\textbf{AwPCSD}} &
  \cellcolor[HTML]{DDFFDD}Proposed method &
  \cellcolor[HTML]{DDFFDD}\textbf{13} &
  \cellcolor[HTML]{DDFFDD}\textbf{23} &
  \cellcolor[HTML]{DDFFDD}\textbf{1340} &
  \cellcolor[HTML]{DDFFDD}\textbf{-} &
  \cellcolor[HTML]{DDFFDD}\textbf{-} &
  \cellcolor[HTML]{DDFFDD}\textbf{0.913} &
  \cellcolor[HTML]{DDFFDD}\textbf{0.972} &
  \cellcolor[HTML]{DDFFDD}\textbf{-} &
  \cellcolor[HTML]{DDFFDD}\textbf{1.512} \\ \hline
 &
   &
  Manual setting &
  12 &
  48 &
  1320 &
  0.0900 &
  0.993 &
  0.973 &
  0.971 &
  0.960 &
  - \\
 &
   &
  \cellcolor[HTML]{E8E8E8}CSA &
  \cellcolor[HTML]{E8E8E8}14 &
  \cellcolor[HTML]{E8E8E8}50 &
  \cellcolor[HTML]{E8E8E8}1000 &
  \cellcolor[HTML]{E8E8E8}0.0100 &
  \cellcolor[HTML]{E8E8E8}0.984 &
  \cellcolor[HTML]{E8E8E8}0.990 &
  \cellcolor[HTML]{E8E8E8}0.971 &
  \cellcolor[HTML]{E8E8E8}0.960 &
  \cellcolor[HTML]{E8E8E8}- \\
\multirow{-3}{*}{\textbf{\begin{tabular}[c]{@{}c@{}}\ref{ss:brain}\\ Results on\\ Brain phantom\end{tabular}}} &
  \multirow{-3}{*}{\textbf{PICCS}} &
  \cellcolor[HTML]{DDFFDD}Proposed method &
  \cellcolor[HTML]{DDFFDD}\textbf{20} &
  \cellcolor[HTML]{DDFFDD}\textbf{20} &
  \cellcolor[HTML]{DDFFDD}\textbf{1460} &
  \cellcolor[HTML]{DDFFDD}\textbf{0.0300} &
  \cellcolor[HTML]{DDFFDD}\textbf{0.998} &
  \cellcolor[HTML]{DDFFDD}\textbf{0.999} &
  \cellcolor[HTML]{DDFFDD}\textbf{0.970} &
  \cellcolor[HTML]{DDFFDD}\textbf{0.960} &
  \cellcolor[HTML]{DDFFDD}\textbf{-}
\end{tabular}
}
\end{table}
\end{landscape}

\end{document}